\documentclass{article}
\usepackage{aaai22}

\usepackage{times}  
\usepackage{helvet}  
\usepackage{courier}  
\usepackage[hyphens]{url}  
\usepackage{graphicx} 
\usepackage{natbib}  
\usepackage{caption} 
\DeclareCaptionStyle{ruled}{labelfont=normalfont,labelsep=colon,strut=off} 
\usepackage{booktabs}       
\usepackage{amsfonts,amsmath}       
\usepackage{amssymb}
\usepackage{amsthm}
\usepackage{nicefrac}       
\usepackage{microtype}      
\usepackage{xcolor}         
\usepackage{multirow}
\usepackage{graphicx}
\usepackage{tikz}

\usepackage{amsmath,amsfonts,bm}

\def\figref#1{figure~\ref{#1}}

\def\eqref#1{equation~\ref{#1}}

\def\1{\bm{1}}

\def\rvb{{\mathbf{b}}}

\def\rvw{{\mathbf{w}}}
\def\rvx{{\mathbf{x}}}

\def\rvz{{\mathbf{z}}}

\def\vzero{{\bm{0}}}
\def\vone{{\bm{1}}}

\def\mA{{\bm{A}}}

\DeclareMathAlphabet{\mathsfit}{\encodingdefault}{\sfdefault}{m}{sl}
\SetMathAlphabet{\mathsfit}{bold}{\encodingdefault}{\sfdefault}{bx}{n}

\def\gE{{\mathcal{E}}}

\def\gG{{\mathcal{G}}}

\def\gR{{\mathcal{R}}}

\def\gV{{\mathcal{V}}}

\title{Neuro-Symbolic Inductive Logic Programming with Logical Neural Networks}
\author{Anonymous Authors}
\author{Prithviraj Sen, Breno W. S. R. de Carvalho, Ryan Riegel, Alexander Gray}
\affiliations{IBM Research}

\newcommand{\squishlist}{
\begin{list}{$\bullet$}
{ \setlength{\itemsep}{0pt} \setlength{\parsep}{3pt}
\setlength{\topsep}{3pt} \setlength{\partopsep}{0pt}
\setlength{\leftmargin}{0em} \setlength{\labelwidth}{1em}
\setlength{\labelsep}{0.5em} } }

\newcommand{\squishlisttwo}{
\begin{list}{$\bullet$}
{ \setlength{\itemsep}{0pt} \setlength{\parsep}{0pt}
\setlength{\topsep}{0pt} \setlength{\partopsep}{0pt}
\setlength{\leftmargin}{2em} \setlength{\labelwidth}{1.5em}
\setlength{\labelsep}{0.5em} } }

\newcommand{\squishend}{
\end{list}  }

\newcommand{\eat}[1]{}
\newcommand{\mysecref}[1]{Section \ref{#1}}
\newcommand{\mytabref}[1]{Table \ref{#1}}
\newcommand{\myeqnref}[1]{Equation \ref{#1}}
\newcommand{\myfigref}[1]{{F}igure \ref{#1}}

\newcommand{\myappref}[1]{{A}ppendix \ref{#1}}

\newtheorem{theorem}{Theorem}[section]

\begin{document}

\maketitle

\begin{abstract}
Recent work on neuro-symbolic inductive logic programming has led to promising approaches that can learn explanatory rules from noisy, real-world data. While some proposals approximate logical operators with differentiable operators from fuzzy or real-valued logic that are parameter-free thus diminishing their capacity to fit the data, other approaches are only loosely based on logic making it difficult to interpret the learned ``rules". In this paper, we propose learning rules with the recently proposed logical neural networks (LNN). Compared to others, LNNs offer strong connection to classical Boolean logic thus allowing for precise interpretation of learned rules while harboring parameters that can be trained with gradient-based optimization to effectively fit the data. We extend LNNs to induce rules in first-order logic. Our experiments on standard benchmarking tasks confirm that LNN rules are highly interpretable and can achieve comparable or higher accuracy due to their flexible parameterization.
\end{abstract}

\section{Introduction}

Inductive logic programming (ILP) \citep{muggleton:ilpw96} has been of long-standing interest where the goal is to learn logical rules from labeled data. Since rules are explicitly symbolic, they provide certain advantages over black box models. For instance, learned rules can be inspected, understood and verified forming a convenient means of storing learned knowledge. Consequently, a number of approaches have been proposed to address ILP including, but not limited to, statistical relational learning \citep{getoor:book} and more recently, neuro-symbolic methods.

\par

Since logical operators such as conjunction and disjunction are not differentiable, one issue that most neuro-symbolic ILP techniques have to address is how to learn rules using gradient-based optimization. A popular solution is to employ extensions from fuzzy or real-valued logic that are either differentiable or have subgradients available. For instance, NeuralLP \citep{yang:nips17} substitutes logical conjunction with product $t$-norm ($x \wedge y \equiv xy$) and logic tensor networks \citep{donadello:ijcai17} with \L ukasiewicz $t$-norm ($x \wedge y \equiv \max(0, x+y-1)$). In an interesting experiment, \citet{evans:jair18} show that among the various options, product $t$-norm seems to lead to the best ILP result. This indicates that the learning approach, rather than the user, should be in charge of substituting logical connectives besides learning the rules themselves. Neural logic machine (NLM) \citep{dong:iclr19} achieves this but at the cost of interpretability. More precisely, it models propositional formulae (consisting of conjunctions, disjunctions and/or negations) with multi-layer perceptrons (MLP). Once trained, it may not be possible to interpret NLM as rules since there exists no standard translation from MLP to logic. What is needed is an extension of classical logic with ties strong enough to be amenable to interpretation and can learn not only rules but also the logical connectives using gradient-based optimization.

\par

In this paper, we propose ILP with the recently proposed logical neural nerworks (LNN) \citep{riegel:arxiv20}. Instead of forcing the user to choose a function that mimics a logical connective, LNNs employ constraints to ensure that neurons behave like conjunctions or disjunctions. By decoupling neuron activation from the mechanism to ensure that it behaves like a logical connective, LNNs offer tremendous flexibility in how to parameterize neurons thus ensuring that they fit the data better while maintaining close connections with classical Boolean logic which, in turn, facilitates principled interpretation. We propose first-order extensions of LNNs that can tackle ILP. Since vanilla backpropagation is insufficient for constraint optimization, we propose flexible learning algorithms capable of handling a variety of (linear) inequality and equality constraints. We experiment with diverse benchmarks for ILP including gridworld and knowledge base completion (KBC) that call for learning of different kinds of rules and show how our approach can tackle both effectively. In fact, our KBC results represents a $4$-$16\%$ relative improvement (in terms of mean reciprocal rank) upon the current best rule-based KBC results on popular KBC benchmarks. Additional, we show that joint learning of rules and logical connectives leads to LNN rules that are easier to interpret vs. other neuro-symbolic ILP approaches.

\eat{\par

\noindent {\bf Roadmap}: We review related work in the next section. In \mysecref{sec:lnn}, we describe in detail generalized logical operators in LNNs.
In \mysecref{sec:firstorder}, we describe learning algorithms and extensions to first-order logic. In \mysecref{sec:experiments}, we compare against baselines on standard ILP benchmarks before concluding with \mysecref{sec:conclusion}. }

\section{Related Work}
\label{sec:relatedwork}

$\partial$ILP \citep{evans:jair18} is another neuro-symbolic ILP technique whose main parameter is a tensor with one cell per candidate logic program. Since the number of candidates is exponential in both the number of available predicates and the number of constituent rules in the program, $\partial$ILP's complexity is exponential making it impractical for anything but the smallest learning task. To reign in the complexity, $\partial$ILP asks the user to specify the ILP task using a \emph{template} consisting of two rules each containing a maximum of two predicates. In reality, most neuro-symbolic ILP approaches ask the user to specify a template. NeuralLP's \citep{yang:nips17} template is meant for link prediction in incomplete knowledge bases (sometimes called open path or chain rule) which only includes binary predicates and is of the form $T(X_1, X_n) \leftarrow R_1(X_1,X_2) \wedge R_2(X_2,X_3) \wedge \ldots R_{n-1}(X_{n-1},X_n)$ positing that the head predicate $T(X_1,X_n)$ can be modeled as a path from $X_1$ to $X_n$. NLM \citep{dong:iclr19} is restricted to learning rules where the head and body predicates each contain the same set of  variables. For instance, to model the rule learned by NeuralLP, NLM would first need to add arguments to $R_1$ so that the new predicate contains all variables including $X_3, \ldots X_n$. In contrast, our approach can use more flexible templates that can express programs beyond two rules, allows the use of $n$-ary predicates ($n$ possibly $> 2$), and allows the head to have fewer variables than the body thus going beyond all of the above mentioned approaches.

\par

Neural theorem provers (NTP) \citep{rocktaschel:nips17} generalize the notion of unification by embedding logical constants into a high-dimensional latent space. NTPs can achieve ILP by learning the embedding for the unknown predicate which forms part of the rule, and subsequently comparing with embeddings of known predicates. NTPs can have difficulty scaling to real-world tasks since the decision to unify two constants is no longer Boolean-valued leading to an explosion of proof paths that need to be explored. To improve scalability, recent proposals either greedily choose (GNTP \citep{minervini:aaai20}) or learn to choose (CTP \citep{minervini:icml20}) most promising proof paths. We compare against CTP in \mysecref{sec:experiments}.

Lifted relational neural networks (LRNN) \citep{sourek:icilp17} model conjunctions using (generalized) sigmoid but fix certain parameters to ensure that it behaves similar to \L ukasiewicz $t$-norm. This limits how well LRNN can model the data, which is contrary to our goal as stated in the previous section. While other combinations of logic and neural networks exist, e.g. logic tensor networks \citep{donadello:ijcai17}, RelNN \citep{kazemi:aaai18}, DeepProbLog \citep{manhaeve:corr18}, to the best of our knowledge, none of these learn rules to address ILP.

\section{Generalized Propositional Logic with Logical Neural Networks}
\label{sec:lnn}

Logical neural networks (LNN) \citep{riegel:arxiv20} allow the use of almost any parameterized function as a logical connective. We illustrate how LNNs generalize conjunction ($\wedge$). Let $0$ denote {\tt false} and $1$ denote {\tt true}. Let $a,b \in \{0,1\}$ and $x, y \in [0,1]$ denote Boolean-valued and continuous-valued variables, respectively. While Boolean logic defines the output of $\wedge$ when $x,y$ attain the extremities of their permissible domains (shown in \myfigref{fig:truthtables} (a)), to fully define real-valued logic's $\wedge$ we need to also extend its definition to $x,y \in (0,1)$. Intuitively, the characteristic \emph{shape} of $\wedge$ is to produce a 1) \emph{low} output when \emph{either input} is \emph{low}, and 2) \emph{high} output when \emph{both inputs} are \emph{high}. A simple way to capture \emph{low} vs. \emph{high} is via a user-defined hyperparameter $\alpha \in (\frac{1}{2}, 1]$: $x \in [0, 1-\alpha]$ constitutes \emph{low} and $x \in [\alpha, 1]$ constitutes \emph{high}. \myfigref{fig:truthtables} (b) expresses $\wedge$ for real-valued logic in terms of $\alpha$.

\begin{figure}
\begin{minipage}{0.4\linewidth}
\centerline{
$
\scriptsize
\begin{array}{cc|c}
a & b & a \wedge b\\
\hline
0 & 0 & 0\\
0 & 1 & 0\\
1 & 0 & 0\\
1 & 1 & 1
\end{array}
$}
\centerline{{\scriptsize (a)}}
\end{minipage}
\begin{minipage}{0.6\linewidth}
\centerline{$
\scriptsize
\begin{array}{cc|c}
x & y & x \wedge y\\
\hline
\lbrack 0, 1-\alpha \rbrack & \lbrack 0, 1-\alpha \rbrack & \lbrack 0, 1-\alpha \rbrack \\
\lbrack 0, 1-\alpha \rbrack  & (1-\alpha, 1] & \lbrack 0, 1-\alpha \rbrack \\
(1-\alpha,1] & \lbrack 0, 1-\alpha \rbrack  & \lbrack 0, 1-\alpha \rbrack \\
\lbrack \alpha,1 \rbrack & \lbrack \alpha,1 \rbrack & \lbrack \alpha,1 \rbrack
\end{array}
$}
\centerline{{\scriptsize (b)}}
\end{minipage}
\caption{(a) Truth table for $\wedge$ in Boolean logic, and (b) Shape of $\wedge$ extended to real-valued logic.}
\label{fig:truthtables}
\end{figure}

LNNs propose constraints to enforce the shape of $\wedge$. Let $f: [0,1] \times [0,1] \rightarrow [0,1]$ denote a monotonically increasing function (in both inputs). In other words, $f(x,y^\prime) \geq f(x,y)$ $\forall y^\prime \geq y$ and $f(x^\prime,y) \geq f(x,y)$ $\forall x^\prime \geq x$. In accordance with \figref{fig:truthtables} (b), LNNs enforce the following constraints:
\begin{eqnarray*}
f(x,y) &\leq 1-\alpha,& ~ \forall ~ x,y \in \lbrack 0, 1-\alpha \rbrack\\
f(x,y) &\leq 1-\alpha,& ~ \forall ~ x \in \lbrack 0, 1-\alpha \rbrack, ~ \forall y \in ( 1-\alpha, 1 \rbrack\\
f(x,y) &\leq 1-\alpha,& ~ \forall ~ x \in ( 1-\alpha, 1 \rbrack, ~ \forall y \in \lbrack 0, 1-\alpha \rbrack\\
f(x,y) &\geq \alpha,& ~ \forall ~ x,y \in \lbrack \alpha, 1 \rbrack
\end{eqnarray*}
Since $f$ is monotonically increasing, we can move all constraints to their corresponding extremities and eliminate the first constraint since it is redundant given the second and third.
\begin{equation*}
f(1-\alpha,1) \leq 1-\alpha, ~ f(1,1-\alpha) \leq 1-\alpha, ~ f(\alpha,\alpha) \geq \alpha\\ 
\end{equation*}

\begin{figure*}
\begin{minipage}{0.195\linewidth}
\scriptsize
\includegraphics[width=\linewidth]{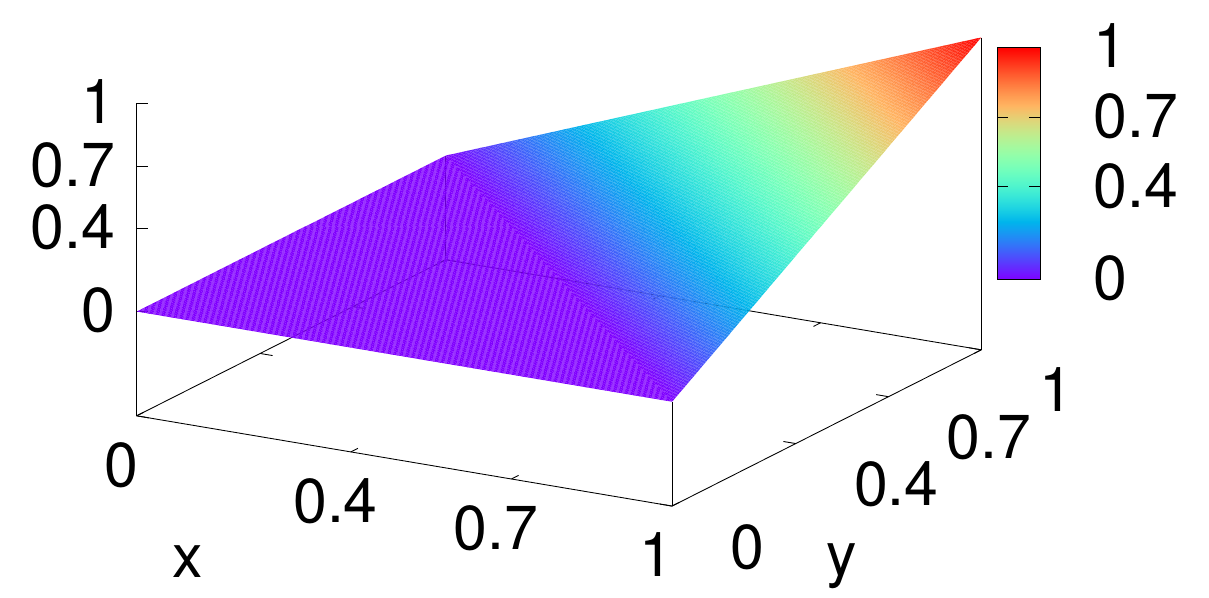}
\centerline{(a)}
\end{minipage}
\begin{minipage}{0.195\linewidth}
\scriptsize
\includegraphics[width=\linewidth]{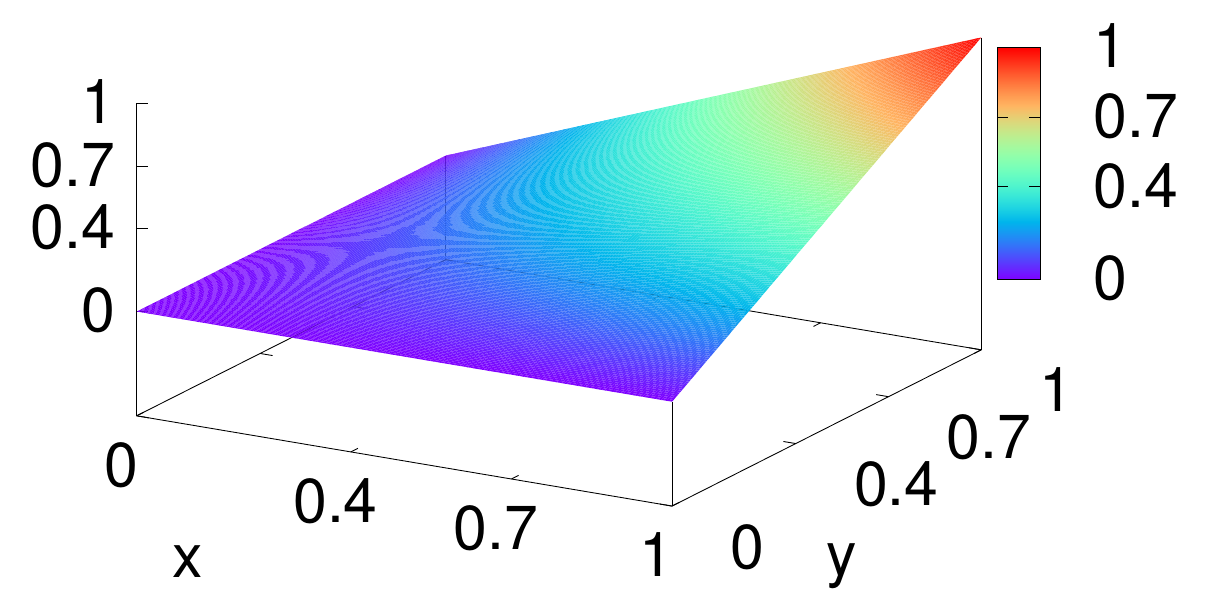}
\centerline{(b)}
\end{minipage}
\begin{minipage}{0.195\linewidth}
\scriptsize
\includegraphics[width=\linewidth]{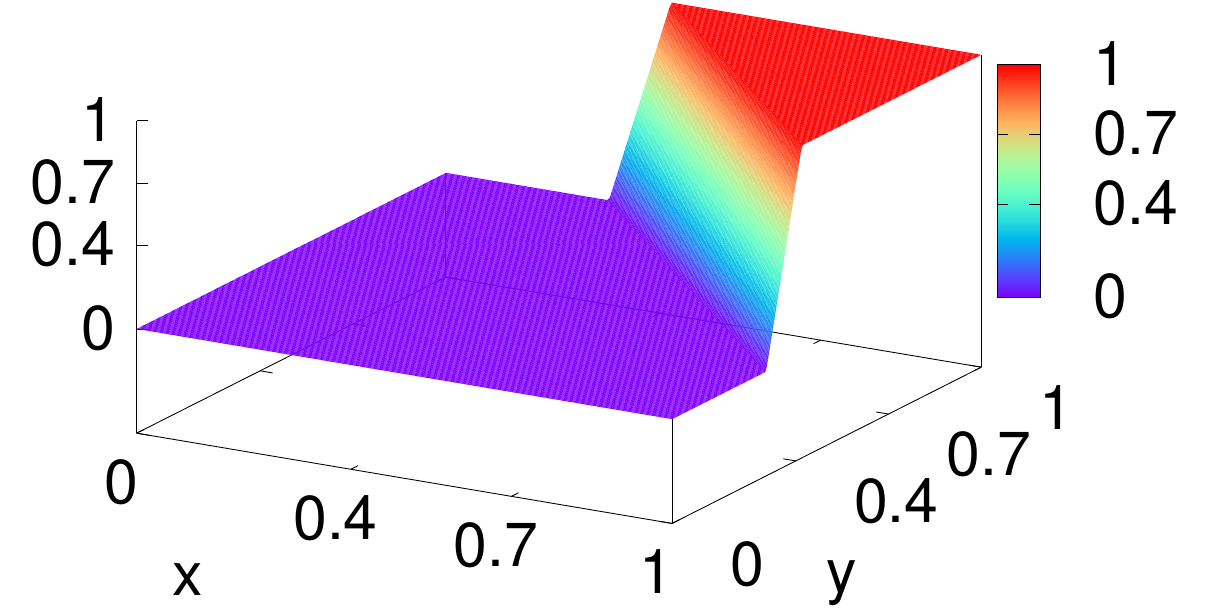}
\centerline{(c)}
\end{minipage}
\begin{minipage}{0.195\linewidth}
\scriptsize
\includegraphics[width=\linewidth]{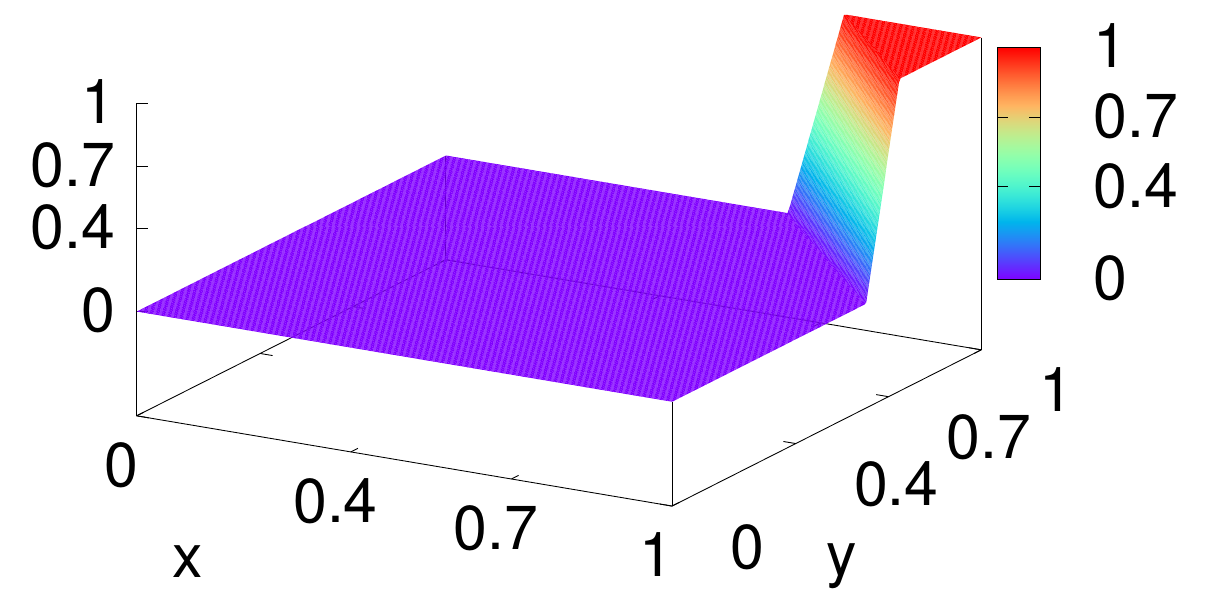}
\centerline{(d)}
\end{minipage}
\begin{minipage}{0.195\linewidth}
\scriptsize
\includegraphics[width=\linewidth]{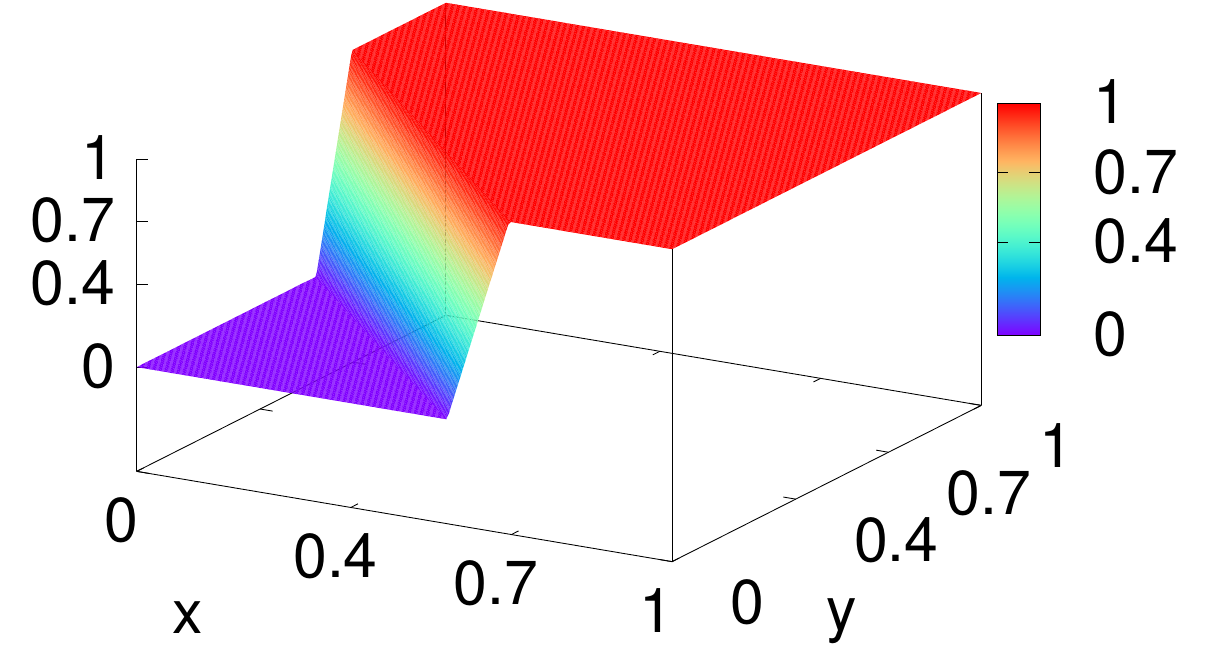}
\centerline{(e)}
\end{minipage}
\caption{(a) Lukasiewicz $t$-norm vs. (b) Product $t$-norm vs. LNN-$\wedge$ with (c) $\alpha=0.7$, (d) $\alpha=0.9$. (e) LNN-$\vee$ ($\alpha=0.7$).}
\label{fig:activations}
\end{figure*}

Further simplifications may be obtained for specific choice of $f$. For inspiration, we look towards \emph{triangular norm} ($t$-norm) \citep{esteva:fuzzysetssyst01} which is defined as a symmetric, associative and non-decreasing function $T: [0,1]^2 \rightarrow [0,1]$ satisfying boundary condition $T(1,x) = x, ~ \forall x \in [0,1]$. Popular $t$-norms include \emph{product} $t$-norm, $xy$, and \emph{\L ukasiewicz} $t$-norm, $\max(0, x+y-1)$. We extend the latter to define LNN-$\wedge$ (other $t$-norms may also be extended similarly):
\begin{eqnarray*}
\lefteqn{\text{LNN-}\!\wedge\!(x,y; \beta, w_1, w_2) =}\\
&&\left\{
\begin{array}{c}
0 \quad \text{ if } \beta - w_1(1-x) - w_2(1-y) < 0\\
1 \quad \text{ if } \beta - w_1(1-x) - w_2(1-y) > 1\\
\beta - w_1(1-x) - w_2(1-y) \quad \text{ otherwise}\\
\end{array}
\right.
\end{eqnarray*} 
where $\beta, w_1, w_2$ denote learnable parameters subject to the following constraints translated from above\footnote{We remove the upper and lower clamps since they do not apply in the active region of the constraints.}:
\begin{eqnarray*}
\text{LNN-}\!\wedge\!(1-\alpha,1;\beta, w_1, w_2) =&& \!\!\!\!\!\!\!\!\!\!\beta - w_1\alpha \leq 1-\alpha \\
\text{LNN-}\!\wedge\!(1,1-\alpha;\beta, w_1, w_2) =&& \!\!\!\!\!\!\!\!\!\!\beta - w_2\alpha \leq 1-\alpha\\
\text{LNN-}\!\wedge\!(\alpha,\alpha;\beta, w_1, w_2) =&& \!\!\!\!\!\!\!\!\!\!\beta - (w_1+w_2)(1-\alpha) \geq \alpha
\end{eqnarray*}
To ensure that LNN-$\wedge$ is monotonically increasing, we also need to enforce non-negativity of $w_1, w_2$. It is easy to extend LNN-$\wedge$ to an $n$-ary conjunction ($n \geq 2$):
\begin{eqnarray}
\nonumber \text{LNN-}\!\wedge\!(\rvx;\beta,\rvw)\equiv&\text{{\tt relu1}}(\beta - \rvw^\top(1-\rvx))&\\
\nonumber \text{subject to:}&\rvw \geq 0, ~ \beta - \alpha \rvw \leq (1-\alpha)\vone&\\
&\beta - (1-\alpha) \vone^\top \rvw \geq \alpha&
\label{eqn:constraints}
\end{eqnarray}
 where {\tt relu1}$(x)$ denotes $\max(0, \min(1, x))$ \citep{krizhevsky:cifar10} and, $\rvx$, $\rvw$ and $\vone$ denote vectors of continuous-valued inputs, weights, and $1$s, respectively.
 
Note that, \myfigref{fig:truthtables} (b) does not enforce constraints on $1-\alpha < x,y < \alpha$. Essentially, $\alpha$ acts as a tunable knob that controls the size of this unconstrained region where we can learn LNN operators without impedance which is an arguably better approach than choosing a parameter-less $t$-norm that would arbitrarily interpolate from Boolean-logic's $\wedge$ to real-valued logic's $\wedge$. \myfigref{fig:activations} illustrates how \L ukasiewicz (\myfigref{fig:activations} (a)) and product (\myfigref{fig:activations} (b)) $t$-norms differ from LNN-$\wedge$ learned by fitting to the four rows in \myfigref{fig:truthtables} (a)'s truth-table. Even pictorially, LNN-$\wedge$ looks distinctly conjunction-like, i.e., when either $x$ or $y$ is low it produces a value close to $0$ while rising quickly to $1$ when both $x,y$ are high. When $\alpha=0.9$ (\myfigref{fig:activations} (d)), the region devoid of constraints is larger than at $\alpha=0.7$ (\myfigref{fig:activations} (c)) ($\because [0.1,0.9] \supset [0.3,0.7]$), so the curve can rise later to provide a better fit . In contrast, \L ukasiewicz $t$-norm remains at $0$ until the $x+y=1$ line, post which it increases linearly. Product $t$-norm is similar, adding a slight, upward curve. 

Other propositional logic operators include negation ($\neg$) and disjunction ($\vee$). LNN negation is given by $1-\rvx$ and LNN disjunction, LNN-$\vee$, is defined in terms of LNN-$\wedge$:
\begin{equation*}
\text{LNN-}\!\vee\!(\rvx;\beta, \rvw) = 1 - \text{LNN-}\!\wedge\!(1-\rvx;\beta,\rvw)
\end{equation*}
where constraints defined in \myeqnref{eqn:constraints} apply. \myfigref{fig:activations} (e) pictorially depicts LNN-$\vee$ (with $\alpha = 0.7$). In contrast to \myfigref{fig:activations} (c), it clearly shows how maximum output is achieved for smaller values of $x, y$, as a disjunction operator should.

\begin{figure*}
\scriptsize
\begin{minipage}{0.1\linewidth}
$\begin{array}{c|cc}
A & X & Y\\
\hline
a_1 & 1 & 2\\
a_2 & 1 & 5\\
\end{array}$
\\
\vspace{4pt}
\\
$\begin{array}{c|cc}
B & X & Y\\
\hline
b_1 & 1 & 2
\end{array}$
\\
\vspace{4pt}
\\
\centerline{
$\begin{array}{c|cc}
C & W & Z\\
\hline
c_1 & 2 & 5
\end{array}$}
\end{minipage}
\begin{minipage}{0.27\linewidth}
\begin{align*}
R(X, Z) &\leftarrow \underline{P}(X,Y) \wedge \underline{Q}(Y,Z)\\
& P \in \{A,B\}, Q \in \{C\}
\end{align*}
\vspace{-10pt}\\
\centerline{\begin{tikzpicture}[>=latex]
\node (r) at (2,0.75) [label=below:{$\wedge$}] {$R(X,Z)$};
\node (p) at (1,0) {$P(X,Y)$};
\node at (1,-0.3) {$P \in \{A,B\}$};
\node (q) at (3,0) {$Q(Y,Z)$};
\node at (3,-0.3) {$Q \in \{C\}$};
\draw[->] (p) -- (r);
\draw[->] (q) -- (r);
\end{tikzpicture}}
\end{minipage}
\begin{minipage}{0.24\linewidth}
\centerline{\begin{tikzpicture}[>=latex]
\node (s) at (3,1.5) [label={below:{$\vee$}}] {$S(X,Z)$};
\node (r) at (2,0.75) [label=below:{$\wedge$}] {$R(X,Z)$};
\node (p) at (1,0) {$P(X,Y)$};
\node at (1,-0.3) {$P \in \{A,B\}$};
\node (q) at (3,0) {$Q(Y,Z)$};
\node at (3,-0.3) {$Q \in \{C\}$};
\node (o) at (4,0.75) {$O(X,Z)$};
\node at (4,0.45) {$O \in \{A,B\}$};
\draw[->] (p) -- (r);
\draw[->] (q) -- (r);
\draw[->] (r) -- (s);
\draw[->] (o) -- (s);
\end{tikzpicture}}
\end{minipage}
\begin{minipage}{0.4\linewidth}
\centerline{$\begin{array}{c|cc}
P & X & Y\\
\hline
p_1 & 1 & 2\\
p_2 & 1 & 5\\
\end{array}$
$\begin{array}{c|cc}
Q & Y & Z\\
\hline
q_1 & 2 & 5
\end{array}$
$\begin{array}{c|cc}
O & X & Z\\
\hline
o_1 & 1 & 2\\
o_2 & 1 & 5\\
\end{array}$}
\vspace{4pt}
\centerline{
$\begin{array}{c|ccc}
PQ & X & Y & Z\\
\hline
pq_1 & 1 & 2 & 5\\
\end{array}$
\hspace{4pt}
$\begin{array}{c|cc}
R & X & Z\\
\hline
r_1 & 1 & 5\\
\end{array}$}
\vspace{4pt}
\centerline{
$\begin{array}{c|cc}
S & X & Z\\
\hline
s_1 & 1 & 5\\
s_2 & 1 & 2\\
\end{array}$}
\end{minipage}
\begin{minipage}{0.1\linewidth}
\scriptsize
\centerline{(a)}
\end{minipage}
\begin{minipage}{0.27\linewidth}
\scriptsize
\centerline{(b)}
\end{minipage}
\begin{minipage}{0.24\linewidth}
\scriptsize
\centerline{(c)}
\end{minipage}
\begin{minipage}{0.4\linewidth}
\scriptsize
\centerline{(d)}
\end{minipage}
\caption{(a) A toy KB. (b) An example rule template (top) and its tree form (bottom). (c) A more complex program template. (d) Generated facts for our running example.}
\label{fig:example}
\end{figure*}
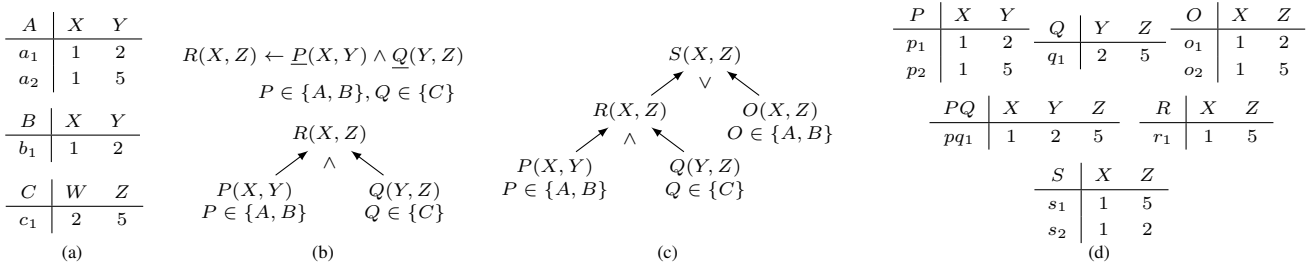

\section{Learning First-Order LNNs}
\label{sec:firstorder}

Following previous work \citep{yang:nips17,evans:jair18,dong:iclr19}, we also utilize program \emph{templates} expressed in higher-order logic to be fed by the user to guide the learning in the right direction. Our definition of a program template draws inspiration from meta-interpretive learning \citep{muggleton:ml14}. In contrast to previous work on neuro-symbolic AI however, our definition of a program template is more general and includes as special cases the templates utilized by \citeauthor{evans:jair18} (considers only rules whose body contains up to 2 predicates), \citeauthor{yang:nips17} (considers only binary predicates) and \citeauthor{dong:iclr19} (considers only rules whose head includes all variables contained in the body). After introducing our logic program template, we then describe how to combine it with data to construct a neural network that may then be trained to learn the logic program of interest.

\par

Let $\text{pred}(X_1, \ldots X_n)$ denote an $n$-ary \emph{predicate} which returns {\tt true} ($1$) or {\tt false} ($0$) for every possible joint assignment of $X_1, \ldots X_n$ to constants in the knowledge base. The main construct in first-order logic (FOL) is a rule or \emph{clause}:
$$
h \leftarrow b_1 \wedge b_2 \wedge \ldots b_m
$$
where $b_1, \ldots b_m$ denote predicates in its \emph{body} and $h$ denotes the \emph{head} predicate. \emph{If} the conjunction of all predicates in the body is true \emph{then} the head is also true. The head $h$ in a clause may contain fewer logical variables than the body $b_1, \ldots b_m$ which means that there must exist an assignment to the missing variables in the head for it to hold. More precisely, if $\mathbf{B} = b_1, \ldots b_m$ denotes the body and is defined over logical variables $\mathbf{Y}$ then $h(\mathbf{X})$, such that $\mathbf{X} \subseteq \mathbf{Y}$, is {\tt true} if $\exists \mathbf{Y} \setminus \mathbf{X}: \mathbf{B}(\mathbf{Y})$ is {\tt true}. Assignments that lead to the predicate being true are also called \emph{facts}.

\par

\myfigref{fig:example} (a) introduces a toy knowledge base (KB) which will serve as a running example. The KB contains three binary predicates, each containing their respective facts along with a unique identifier for easy reference. Thus, $A(X,Y)$'s facts are $A(1, 2)$ (denoted $a_1$) and $A(1,5)$ (denoted $a_2$). \myfigref{fig:example} (b) shows a template for learning $R(X,Z)$ with three crucial pieces of information: 1) the first predicate in its body $P$ is binary and while we do not know the identity of this predicate we know its domain $\text{Dom}(P)$ is $\{A, B\}$, 2) to keep the example simple, the second predicate in the body $Q$ has a singleton domain ($\{C\}$), and lastly 3) the second argument in the first predicate should be equal to the first argument in the second predicate indicated by repeated use of $Y$. \myfigref{fig:example} (b) (bottom) expresses the same template as a tree where $P(X,Y)$ and $Q(Y,Z)$ are annotated with their respective domains forming children of $R(X,Z)$ whose label $\wedge$ indicates that the predicates in its body are to be conjuncted together.

\par

\myfigref{fig:example} (c) shows a more complex template that disjuncts $R(X,Z)$ with $O(X,Z)$, such that $\text{Dom}(O) = \{A,B\}$, to produce $S(X,Z)$. \myfigref{fig:example} (d) shows \emph{generated} facts (with unknown truth-values) that can be possibly produced when this template is combined with the KB from \myfigref{fig:example} (a). $P, Q$ and $O$ contain the union of all facts included in the predicates in their respective domains. Since $p_1$ and $q_1$ are the only two facts that agree on value of $Y$, $PQ$, the predicate produced by the body of $R$, contains only one generated fact. $R$ is obtained by dropping $Y$, which is then unioned with $O$ to produce $S$. By comparing generated facts in $S$ with labels in the training data, it is possible to learn a full program which in this case constitutes learning: 1) which predicates to replace $P, Q$ and $O$ with, and 2) the logical connectives, LNN-$\wedge$ and LNN-$\vee$, used to model $R$ and $S$ with, respectively. We next state a more formal problem definition.

\par

Let $\mathcal{T} = (\mathcal{V}, \mathcal{E}, \mathcal{L})$ denote a tree-structured \emph{program template} where $\mathcal{V}$ denotes the set of nodes, $\mathcal{E}$ denotes the set of edges and $\mathcal{L}$ denotes a mapping from $\mathcal{V}$ to node labels. $\mathcal{L}$ maps $\mathcal{T}$'s leaves to the corresponding domain of predicates in the KB. In the worst case, the domain can be the subset of predicates in the KB that agrees with the arity of the leaf. $\mathcal{L}$ maps internal nodes to a logical operator $\{\wedge, \vee, \neg\}$. The ILP task can then be stated as, given $\mathcal{T}$, a knowledge base KB, and truth-values corresponding to generated facts in the root of $\mathcal{T}$, to learn all logical connectives involved along with selecting predicates for each leaf in $\mathcal{T}$.

\par

\eat{\myfigref{fig:example} (d) shows that $s_1, s_2$ are the generated facts in $S(X,Z)$, the root of the template in \myfigref{fig:example} (c). Our goal is to achieve ILP given truth-values for such facts, e.g., the labeled data might state that truth-value for $s_1$ is $1$ and $s_2$ is $0$. To predict the truth-values of $s_1, s_2$, we need to consider their lineage comprising other facts. For instance, lineage of $s_1$ contains $r_1$ and $o_2$. In the sequel, we describe how to compute $\psi(v)$, denoting the truth-value associated with generated fact $v$, from truth-values of other facts in $v$'s lineage. We represent the whole computation as a neural network that can then be learned end-to-end using backpropagation. In the ensuing discussion, let $\mathbf{V}(\mathbf{X})$ denote some node in $\mathcal{T}$ whose predicate is defined over variables $\mathbf{X}$, $\mathbf{V}(\mathbf{x})$ denote a fact obtained by the subtitution $\mathbf{X}=\mathbf{x}$, $\mathcal{F}(\mathbf{V})$ denote all facts of predicate $\mathbf{V}$ and $\mathcal{N}(\mathbf{V})$ denote children of $\mathbf{V}$ in $\mathcal{T}$.}

In the remainder of this section, we describe how to achieve the above ILP task given ground truth values for generated facts belonging to the root of the template. Let $\psi(v)$ denote the truth value associated with (generated) fact $v$. Our strategy is to build a neural network that connects the truth values of generated facts from the root of the template to other (generated) facts in its lineage right down to the facts in the base KB whose truth values are defined to be $1$. The whole neural network can then be trained end-to-end using backpropagation. Let $\mathbf{V}(\mathbf{X})$ denote any node in $\mathcal{T}$ whose predicate is defined over variables $\mathbf{X}$ and whose children in $\mathcal{T}$ is denoted by $\mathcal{N}(\mathbf{V})$. Also, let $\mathbf{V}(\mathbf{x})$ denote a fact obtained by the substitution $\mathbf{X}=\mathbf{x}$ and $\mathcal{F}(\mathbf{V})$ denote all facts of $\mathbf{V}$.

\subsection{Combining Base Facts}
\eat{
\begin{figure*}
\centerline{
\begin{minipage}{0.3\linewidth}
\includegraphics[width=0.99\linewidth]{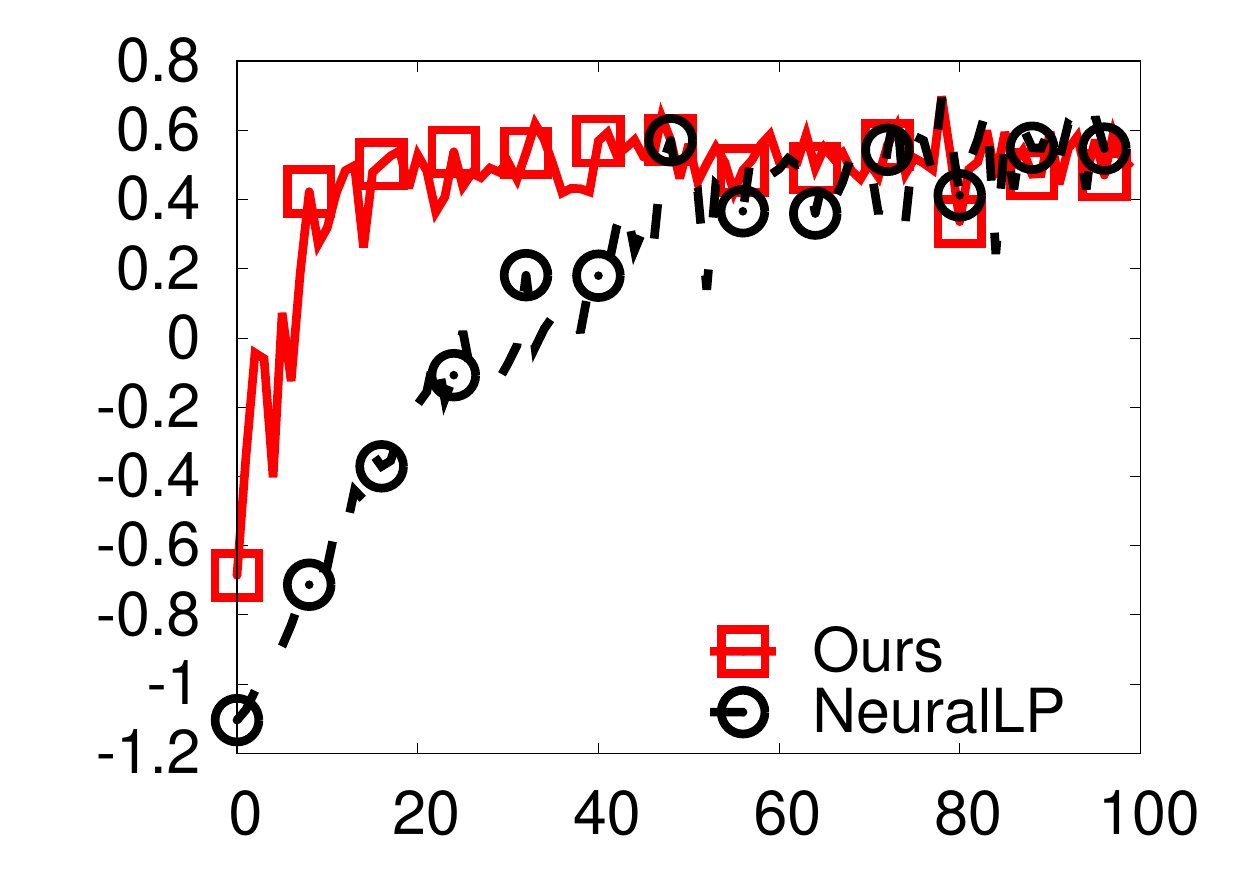}
\end{minipage}
\begin{minipage}{0.4\linewidth}
\scriptsize
\begin{center}
\begin{tikzpicture}[=>latex]
\node (noobssouth) at (-0.2,0) [draw=blue!80,fill=blue!20] {$\neg${\tt HasObstacleSouth}$(X,Y)$};
\node[label=left:{$(1.056)$}] (pred1) at (-0.2,1) [draw=red!80,fill=red!20] {LNN-{\tt pred}};
\node (tgtsouth) at (3.2,0) [draw=blue!80,fill=blue!20] {{\tt HasTargetSouth}$(X,Y)$};
\node[label=right:{$(1.005)$}] (pred2) at (3.2,1) [draw=red!80,fill=red!20] {LNN-{\tt pred}};
\node[label=above:{$(1.791)$}] (and) at (1.5,2) [draw=gray!80,fill=lightgray!20]{LNN-$\wedge$};
\draw[->,thick] (pred1) -- (and) node [midway,left=2pt] {$2.239$};
\draw[->,thick] (pred2) -- (and) node [midway,right=2pt] {$2.322$};
\draw[->,thick] (noobssouth) -- (pred1) node [midway,left=2pt] {$1.077$};
\draw[->,thick] (tgtsouth) -- (pred2) node[midway,right=2pt]  {$1.052$};
\node at (0,2.5) {\vspace{0pt}};
\node at (0,-0.5) {\vspace{0pt}};
\end{tikzpicture}
\end{center}
\end{minipage}
\begin{minipage}{0.3\linewidth}
\includegraphics[width=0.99\linewidth]{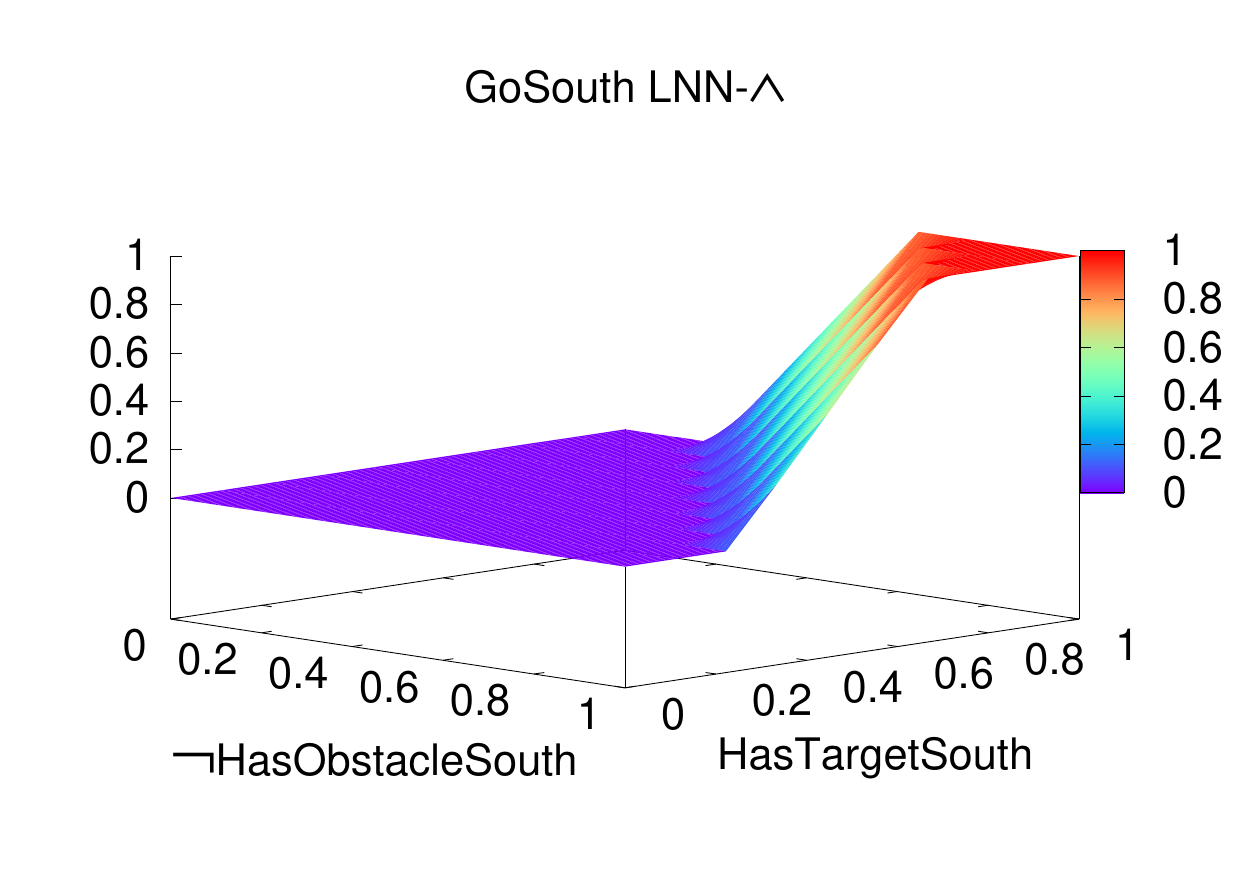}
\end{minipage}}
\centerline{
\begin{minipage}{0.3\linewidth}
\scriptsize
\centerline{(a)}
\end{minipage}
\begin{minipage}{0.4\linewidth}
\scriptsize
\centerline{(b)}
\end{minipage}
\begin{minipage}{0.3\linewidth}
\scriptsize
\centerline{(c)}
\end{minipage}
}
\caption{Gridworld: (a) Avg. Rewards vs. Training Grids. (b) LNN Rule and (c) LNN-$\wedge$ for GoSouth(X,Y).}
\label{fig:gridworld}
\end{figure*}}

Let $\mathbf{V}(\mathbf{X})$ denote a leaf in $\mathcal{T}$ with domain $\mathcal{L}(\mathbf{V})$ then $\mathcal{F}(\mathbf{V})$ is given by  $\bigcup_{\mathbf{P} \in \mathcal{L}(\mathbf{V})} \mathcal{F}(\mathbf{P})$. Computing $\psi(\mathbf{V}(\mathbf{X}=\mathbf{x}))$ corresponding to $\mathbf{X} = \mathbf{x}$ requires truth values of all facts corresponding to the same substitution. We provide two options that associate parameters with predicates in $\mathcal{L}(\mathbf{V})$: 1) \emph{attention} \citep{yang:nips17} and 2) our proposed LNN-{\tt pred} operator:
\begin{align*}
\text{1)}\quad\psi(\mathbf{V}(\mathbf{x})) =& \sum_{\mathbf{P} \in \mathcal{L}(\mathbf{V})} w_{\mathbf{P}} \psi(\mathbf{P}(\mathbf{x}))\\
\text{2)}\quad\psi(\mathbf{V}(\mathbf{x})) =& 1- \mathtt{relu1}\left(\beta - \sum_{\mathbf{P} \in \mathcal{L}(\mathbf{V})} w_{\mathbf{P}} \psi(\mathbf{P}(\mathbf{x}))\right)
\end{align*}
where $\beta$, $w_{\mathbf{P}}$ denote learnable parameters. One issue with attention is that it may lack sparsity assigning a majority of predicates in $\mathcal{L}(\mathbf{V})$ non-zero weights thus hampering interpretability. To address this, we propose an alternate parameterization, LNN-{\tt pred}, that is a simpler version of LNN-$\vee$ and lacks all constraints except for non-negativity of $w_{\mathbf{P}}$ (since we do not require disjunctive semantics). As an example, the lower left corner of \myfigref{fig:nnexample} shows how to compute $\psi(p_1)$ from $\psi(a_1), \psi(b_1)$ since $a_1,b_1$ form the lineage for $p_1$ (\myfigref{fig:example}). Here, $\beta_1, \rvw_1$ denote LNN-{\tt pred}'s parameters.

\subsection{Combining Facts with Conjunction}

We handle conjunctions in two steps: 1) first construct the result of the body of the clause, and then 2) construct the head, dropping variables if needed. Let $\mathbf{V}(\mathbf{Y})$ in $\mathcal{T}$ be such that $\mathcal{L}(\mathbf{V}) = \wedge$ and $\mathcal{N}(\mathbf{V})$ denote its children. Also, let $\mathbf{I}(\mathbf{X})$ denote the intermediate predicate produced from the body of $\mathbf{V}(\mathbf{Y})$ potentially containing additional variables such that $\mathbf{X} \supseteq \mathbf{Y}$. We use LNN-$\wedge$ to compute $\psi$ of $\mathbf{I}(\mathbf{x})$:
\begin{align*}
\psi(\mathbf{I}(\mathbf{x})) = \text{{\tt relu1}}\left(\beta - \sum_{\mathbf{P} \in \mathcal{N}(\mathbf{V})} w_\mathbf{P}(1-\psi(\mathbf{P}(\mathbf{x}_{\text{var}(\mathbf{P})})))\right)
\end{align*}
where $\text{var}(\mathbf{P})$ denotes predicate $\mathbf{P}$'s arguments and $\mathbf{x}_{\text{var}}$ denotes the substitution restricted to variables in var. When $\mathbf{X} \supset \mathbf{Y}$, multiple facts from $\mathcal{F}(\mathbf{I})$ may combine to produce a fact in $\mathbf{V}(\mathbf{Y})$ and we use {\tt maxout} for this:
\begin{equation*}
\psi(\mathbf{V}(\mathbf{y})) = \text{{\tt maxout}}(\{\psi(\mathbf{I}(\mathbf{x})) ~|~ \mathbf{x}_{\mathbf{Y}} = \mathbf{y}, \forall \mathbf{I}(\mathbf{x}) \in \mathcal{F}(\mathbf{I})\})
\end{equation*}
where $\text{{\tt maxout}}(\{x_1, \ldots\})$ \citep{goodfellow:icml13} returns the maximum of the set. \myfigref{fig:nnexample} shows how $\psi(pq_1)$ is computed from $\psi(p_1),\psi(q_1)$ where $\beta_2, \rvw_2$ denotes LNN-$\wedge$'s parameters. Since $pq_1$ is the only intermediate fact leading to $r_1$, we do not need {\tt maxout} in this case. However, if that was not the case, \myfigref{fig:nnexample} shows where {\tt maxout} would appear.

\subsection{Combining Facts with Disjunction}

\begin{figure}
\scriptsize
\begin{tikzpicture}[>=latex]
\node (a1) at (-1.1,0) {$1 (a_1)$};
\node (b1) at (-0.1,0) {$1 (b_1)$};

\node (beta1_hat) at (-2.7,1.45) [rectangle, draw, minimum width=5.5mm,fill=lightgray!50] {$\widehat{\beta_1}$};
\node (w1_hat) at (-2.7,0.75) [rectangle, draw, minimum height=1cm, minimum width=5.5mm,draw,fill=lightgray!50] {$\widehat{\rvw_1}$};
\node (w1) at (-1.5,0.75) [rectangle, draw, minimum height=1cm, minimum width=5.5mm,draw,fill=lightgray!50] {$\rvw_1$};
\node (beta1) at (-1.5,1.45) [rectangle, draw, minimum width=5.5mm,fill=lightgray!50] {$\beta_1$};

\draw[->] (beta1_hat) -- (beta1) node[midway, above, align=center] {{\tt relu}};
\draw[->] (w1_hat) -- (w1) node[midway, above, align=center] {{\tt relu}};

\node (dot1) at (-0.6,0.75) [circle,draw,minimum size=4mm,fill=blue!20]{};
\draw[->] (w1) -- (dot1);
\draw[->] (a1) -- (dot1);
\draw[->] (b1) -- (dot1);

\node at (0.2,0.75) {$[1,1]\rvw_1$};
\node (act1) at (-0.6,1.45) [circle,draw,minimum size=4mm,fill=blue!20]{};
\node at (-0.2,1.45) {$-$};
\draw[->] (dot1) -- (act1);
\draw[->] (beta1) -- (act1);

\draw[-,dashed] (-3.05,0.2) -- (-1.15,0.2) -- (-1.15,1.75) -- (-3.05,1.75) -- (-3.05,0.2);
\node at (-2.4,0) {{\bf LNN-{\tt pred}}};

\node (p1) at (0.3,2.25) {$p_1$};
\draw[->] (act1) -- (-0.6,1.85) -- (0.3,1.85) -- (p1);
\node at (-0.7,2) {1-{\tt relu-1}$(\cdot)$};

\node (c) at (1.1,1.55) {$1 (c_1)$};
\node (q1) at (1.1,2.25) {$q_1$};
\draw[->] (c) -- (q1);

\node (mu1_tilde) at (-1.09,5.3) [rectangle, draw, minimum height=8mm, minimum width=3mm,fill=green!20] {$\widehat{\mu_2}$};
\node (mu1) at (-1.09,4.1) [rectangle, draw, minimum height=8mm, minimum width=3mm] {$\mu_2$};
\node (gamma1) at (-1.8,4.1) [rectangle, draw, minimum height=8mm, minimum width=8mm,fill=red!20] {$\Gamma_2$};
\node (gen11) at (-1.25,3) [rectangle, draw, minimum height=8mm, minimum width=3mm] {$\Gamma_2 \mu_2$};
\draw[->] (mu1_tilde) --  (mu1) node[midway, left, align=center] {{\tt relu}};
\draw[->] (mu1.south) --  ([xshift=4.5pt]gen11.north);
\draw[->] ([xshift=-2pt]gamma1.south) -- ([xshift=-2pt]gamma1.south |- gen11) -- (gen11);

\node (gen21) at (-2.6,3) [rectangle, draw, minimum height=8mm, minimum width=3mm] {$\Upsilon_2 \lambda_2$};
\node (lambda1) at (-2.78,4.1) [rectangle, draw, minimum height=8mm, minimum width=3mm] {$\lambda_2$};
\draw[->] (lambda1) -- ([xshift=-5pt]gen21.north);
\node (lambda1_tilde) at (-2.78,5.3) [rectangle, draw,fill=green!20, minimum height=8mm, minimum width=3mm] {$\widehat{\lambda_2}$};
\draw[->] (lambda1_tilde) -- (lambda1) node[midway, left, align=center] {$\sigma$};
\node (nu1) at (-2.05,5.3) [rectangle, draw, minimum height=8mm, minimum width=8mm,,fill=red!20] {$\Upsilon_2$};
\draw[->] ([xshift=-8pt]nu1.south) -- ([xshift=7.5pt]gen21.north);

\node (w2) at (-0.2,3) [rectangle, draw, minimum height=1cm, minimum width=5.5mm,draw,fill=lightgray!50] {$\rvw_2$};
\node (beta2) at (-0.2,3.7) [rectangle, draw, minimum width=5.5mm,fill=lightgray!50] {$\beta_2$};
\draw[->] (gen11) -- (w2);
\draw[->] (gen21.south) -- (gen21.south |- w2.south west) -- (w2.south west);
\node at (-0.67,2.75) {+};

\node (dot2) at (0.7,3) [circle,draw,minimum size=4mm,fill=blue!20]{};
\draw[->] (w2) -- (dot2);
\draw[->] (p1) -- (dot2);
\draw[->] (q1) -- (dot2);

\draw[-,dashed] (-3.1,2.4) -- (-3.1,5.75) -- (-0.75,5.75) -- (-0.75,3.95) -- (0.15,3.95) -- (0.15,2.4) -- (-3.1,2.4);
\node at (-2.65,2.25) {{\bf LNN-$\wedge$}};

\node at (2.1,2.8) {$\rvw_2^\top (1-[p_1,q_1]^\top)$};
\node (act2) at (0.7,3.7) [circle,draw,minimum size=4mm,fill=blue!20]{};
\node at (0.4,3.5) {$-$};
\draw[->] (dot2) -- (act2);
\draw[->] (beta2) -- (act2);

\node (pq1) at (0.7,4.5) {$pq_1$};
\draw[->] (act2) -- (pq1) node[midway, left, align=center] {{\tt relu-1}};
\node (r1) at (0.7,5.3) {$r_1$};
\draw[->] (pq1) -- (r1) node[midway, left, align=center] {{\tt maxout}};

\node (o1) at (1.5,5.3) {$o_2$};
\node (act3) at (2.1,1.45) [circle,draw,minimum size=4mm,fill=blue!20]{};
\node at (1.7,1.45) {$-$};
\draw[->] (act3) -- (2.1,2.1) -- (3.5,2.1) -- (3.5,3.1) -- (1.5,3.1) -- (o1);
\node at (2.8,2.25) {1-{\tt relu-1}$(\cdot)$};
\node (dot3) at (2.1,0.75) [circle,draw,minimum size=4mm,fill=blue!20]{};
\draw[->] (dot3) -- (act3);
\node (a2) at (2.1,0) {$1 (a_2)$};
\draw[->] (a2) -- (dot3);
\node at (1.4,0.75) {$[1,0]\rvw_3$};

\node (beta3_hat) at (4.2,1.45) [rectangle, draw, minimum width=5.5mm,fill=lightgray!50] {$\widehat{\beta_3}$};
\node (w3_hat) at (4.2,0.75) [rectangle, draw, minimum height=1cm, minimum width=5.5mm,draw,fill=lightgray!50] {$\widehat{\rvw_3}$};
\node (w3) at (3,0.75) [rectangle, draw, minimum height=1cm, minimum width=5.5mm,draw,fill=lightgray!50] {$\rvw_3$};
\node (beta3) at (3,1.45) [rectangle, draw, minimum width=5.5mm,fill=lightgray!50] {$\beta_3$};
\draw[->] (beta3_hat) -- (beta3) node[midway, above, align=center] {{\tt relu}};
\draw[->] (w3_hat) -- (w3) node[midway, above, align=center] {{\tt relu}};
\draw[->] (w3) -- (dot3);
\draw[->] (beta3) -- (act3);
\draw[-,dashed] (2.65,0.2) -- (4.58,0.2) -- (4.58,1.75) -- (2.65,1.75) -- (2.65,0.25);
\node at (3.9,0) {{\bf LNN-{\tt pred}}};

\node (act4) at (1.1,6.7) [circle,draw,minimum size=4mm,fill=blue!20]{};
\node (dot4) at (1.1,6) [circle,draw,minimum size=4mm,fill=blue!20]{};
\draw[->] (r1) -- (dot4);
\draw[->] (o1) -- (dot4);
\draw[->] (dot4) -- (act4);
\node at (0.7,6.7) {$-$};
\node at (0.25,6) {$[r_1,o_2]\rvw_4$};
\node (s1) at (1.1,7.4) {$s_1$};
\draw[->] (act4) -- (s1) node[midway, left, align=center]{1-{\tt relu-1}$(\cdot)$};

\node (w4) at (2,6) [rectangle, draw, minimum height=1cm, minimum width=5.5mm,draw,fill=lightgray!50] {$\rvw_4$};
\node (beta4) at (2,6.7) [rectangle, draw, minimum width=5.5mm,fill=lightgray!50] {$\beta_4$};
\draw[->] (w4) -- (dot4);
\draw[->] (beta4) -- (act4);

\node (gen24) at (4.4,6) [rectangle, draw, minimum height=8mm, minimum width=3mm] {$\Upsilon_4 \lambda_4$};
\node (gen14) at (3.05,6) [rectangle, draw, minimum height=8mm, minimum width=3mm] {$\Gamma_4 \mu_4$};
\draw[->] (gen14) -- (w4);
\draw[->] (gen24.north) -- (gen24.north |- w4.north east) -- (w4.north east);
\node at (2.45,6.25) {+};
\node (mu4) at (2.89,4.9) [rectangle, draw, minimum height=8mm, minimum width=3mm] {$\mu_4$};
\node (gamma4) at (3.6,4.9) [rectangle, draw, minimum height=8mm, minimum width=8mm,fill=red!20] {$\Gamma_4$};
\node (lambda4) at (4.58,4.9) [rectangle, draw, minimum height=8mm, minimum width=3mm] {$\lambda_4$};
\draw[->] (mu4.north) --  ([xshift=-4.5pt]gen14.south);
\draw[->] ([xshift=2pt]gamma4.north) -- ([xshift=2pt]gamma4.north |- gen14) -- (gen14);
\node (lambda4_tilde) at (4.58,3.7) [rectangle, draw,fill=green!20, minimum height=8mm, minimum width=3mm] {$\widehat{\lambda_4}$};
\node (mu4_tilde) at (2.89,3.7) [rectangle, draw, minimum height=8mm, minimum width=3mm,fill=green!20] {$\widehat{\mu_4}$};
\node (nu4) at (3.85,3.7) [rectangle, draw, minimum height=8mm, minimum width=8mm,,fill=red!20] {$\Upsilon_4$};
\draw[->] (lambda4) -- ([xshift=5pt]gen24.south);
\draw[->] (lambda4_tilde) -- (lambda4) node[midway, right, align=center] {$\sigma$};
\draw[->] (mu4_tilde) --  (mu4) node[midway, right, align=center] {{\tt relu}};
\draw[->] ([xshift=8pt]nu4.north) -- ([xshift=-7.5pt]gen24.south);
\node at (4.4,3.1) {{\bf LNN-$\vee$}};
\draw[-,dashed] (2.55,3.23) -- (4.9,3.23) -- (4.9,6.95) -- (1.65,6.95) -- (1.65,5.45) -- (2.55,5.45) -- (2.55,3.23);

\end{tikzpicture}
\caption{Neural network constructed for $s_1 \in \mathcal{F}(S)$. $\sigma$ denotes {\tt softmax}.}
\label{fig:nnexample}
\end{figure}
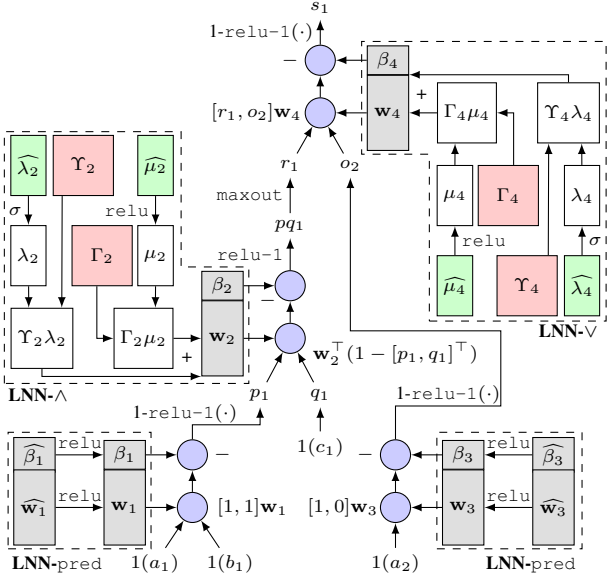

Given $\mathbf{V}(\mathbf{X})$ in $\mathcal{T}$ such that $\mathcal{L}(\mathbf{V}) = \vee$, $\psi(\mathbf{V}(\mathbf{x}))$ can be computed using LNN-$\vee$:
\begin{equation*}
\psi(\mathbf{V}(\mathbf{x})) = 1 - \text{{\tt relu1}}\left(\beta - \sum_{\mathbf{P} \in \mathcal{N}(\mathbf{V})} w_\mathbf{P}\psi(\mathbf{P}(\mathbf{x}))\right)
\end{equation*}
In \myfigref{fig:nnexample} shows how $\psi(s_1)$ is computed from $\psi(r_1)$ and $\psi(o_2)$ where $\beta_r, \rvw_4$ denotes LNN-$\vee$'s parameters.

\subsection{Other Operations and Extensions}
\label{sec:extensions}

Implementing negation is more involved since it requires that we consider assignments that \emph{may} lead to facts in $\mathbf{V}$ but are not present in its child predicate $\mathbf{P}$. Given a universe of all possible assignments $\mathcal{U}$, we express $\psi(\mathbf{V}(\mathbf{x}))$ as:
\begin{equation*}
\psi(\mathbf{V}(\mathbf{x})) = 1 - \psi(\mathbf{P}(\mathbf{x})), ~ \forall \mathbf{x} \in \mathcal{U}
\end{equation*}
where $\psi(\mathbf{P}(\mathbf{x}))$ is defined as $0$ if $\mathbf{P}(\mathbf{x}) \notin \mathcal{F}(\mathbf{P})$.

Note that, any LNN operator introduced for $\mathbf{V}(\mathbf{X})$ is shared across all $\mathbf{V}(\mathbf{x}) \in \mathcal{F}(\mathbf{V})$ since the result of ILP should be agnostic of individual facts. Thus, even though the (sub)network constructed for $\mathbf{V}(\mathbf{x})$ may differ from $\mathbf{V}(\mathbf{x}^\prime)$'s, e.g., $s_2$'s neural network (not shown) is simpler than $s_1$'s, gradient updates flow to the same LNN parameters. For simplicity, we only discussed templates comprising a tree of Horn clauses but the ideas presented here can easily extend to DAG-structured templates and going beyond equality conditions in the body of a clause, e.g., $R(X,Z) \leftarrow P(X,Y) \wedge Q(Y^\prime,Z) \wedge Y > Y^\prime$.

\subsection{Learning Constrained Activations}

So far, we have shown how to construct a neural network from a KB and template comprising parameters of LNN operators but we have not addressed how to enforce constraints on said parameters. More precisely, $\beta_i, \rvw_i, ~ \forall i=1,\ldots 4$ in \myfigref{fig:nnexample} need to satisfy the respective constraints associated with the LNN operators they form parameters for (as described in \mysecref{sec:lnn}). Since backpropagation does not handle constraints, we propose to apply a recently proposed approach to ``fold" in a system of linear constraints as layers into the neural network \citep{frerix:cvprw20}. We note that \citet{riegel:arxiv20} also devise a training algorithm for learning LNN operators but this is tightly connected to a specific kind of LNN operator called tailored activation. For the small systems of inequality constraints introduced by LNN-$\wedge$, LNN-$\vee$ and LNN-{\tt pred}, the approach presented here conveniently allows learning all (constrained) parameters of LNNs using vanilla backpropagation alone.

\par

\citet{frerix:cvprw20} recently showed how to handle a system of linear inequality constraints of the form $\mA \rvz \leq \rvb$ where $\mA$ denotes a matrix containing coefficients in the constraints, $\rvz$ denotes the parameters (in our case, some concatenation of $\beta$ and $\rvw$), and $\rvb$ denotes the constants in the constraints. We begin with the Minkowski-Weyl theorem:
\begin{theorem}
A set $\mathcal{C} = \{\rvz \, | \, \mA \rvz \leq \rvb\}$ is a convex polyhedron if and only if:
\begin{equation*}
\mathcal{C} = \left\{\Upsilon \mu + \Gamma \lambda \, \middle| \, \mu, \lambda \geq \vzero, \vone^\top \lambda = 1\right\}
\end{equation*}
where $\Upsilon$ and $\Gamma$ contain a finite number of rays and vertices, respectively.
\end{theorem}
\noindent which states that there exists a translation from $\mA, \rvb$ to $\Upsilon, \Gamma$ obtained via the \emph{double-description} method \citep{motzkin:tog53}. Assuming we can generate non-negative vectors $\mu, \lambda$ and additionally ensure that $\lambda$ sums to $1$, then one can access a point $\rvz = [\beta, \rvw^\top]^\top$ from the feasible set $\mathcal{C}$ by computing $\Upsilon \mu + \Gamma \lambda$. Sampling vectors $\mu, \lambda$ can be achieved, for instance, by using {\tt relu} \citep{nair:icml10} and {\tt softmax}:

\begin{minipage}{0.35\linewidth}
\begin{equation*}
\mu = \max(0, \widehat{\mu})
\end{equation*}
\end{minipage}
\hfill
\begin{minipage}{0.5\linewidth}
\begin{equation*}
\lambda = \frac{\exp(\widehat{\lambda})}{Z}, ~ Z = \vone^\top\exp(\widehat{\lambda})
\end{equation*}
\end{minipage}
\begin{align*}
[\beta, \rvw^\top]^\top &= \Upsilon\mu + \Gamma\lambda
\end{align*}
Additionally, these operations can be easily included into any neural network as additional layers. For instance, \myfigref{fig:nnexample} contains in dashed boxes the above set of layers needed to generate $\rvw_i, \beta_i, ~ \forall i=1,\ldots 4$. The resulting neural network is self-contained, and can be trained by vanilla backpropagation end-to-end thus ensuring that the learned LNN parameters satisfy their respective constraints.

\section{Experiments}
\label{sec:experiments}

\begin{figure*}
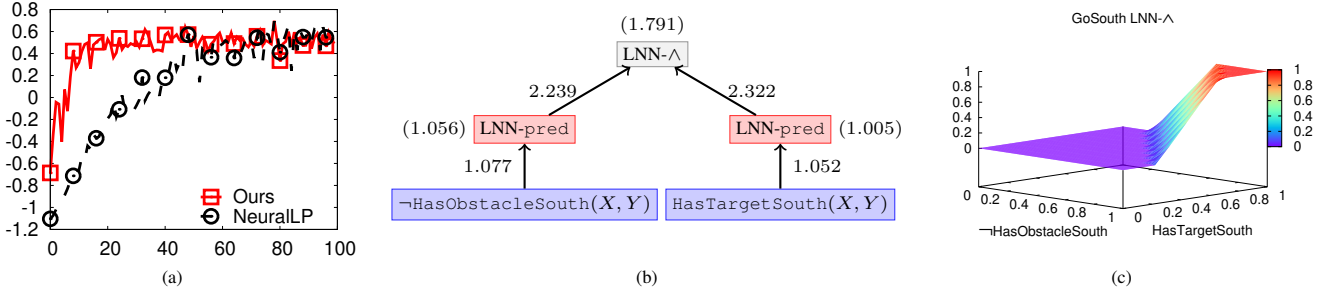

\centerline{
\begin{minipage}{0.3\linewidth}
\includegraphics[width=0.99\linewidth]{rewards}
\end{minipage}
\begin{minipage}{0.4\linewidth}
\scriptsize
\begin{center}
\begin{tikzpicture}[=>latex]
\node (noobssouth) at (-0.2,0) [draw=blue!80,fill=blue!20] {$\neg${\tt HasObstacleSouth}$(X,Y)$};
\node[label=left:{$(1.056)$}] (pred1) at (-0.2,1) [draw=red!80,fill=red!20] {LNN-{\tt pred}};
\node (tgtsouth) at (3.2,0) [draw=blue!80,fill=blue!20] {{\tt HasTargetSouth}$(X,Y)$};
\node[label=right:{$(1.005)$}] (pred2) at (3.2,1) [draw=red!80,fill=red!20] {LNN-{\tt pred}};
\node[label=above:{$(1.791)$}] (and) at (1.5,2) [draw=gray!80,fill=lightgray!20]{LNN-$\wedge$};
\draw[->,thick] (pred1) -- (and) node [midway,left=2pt] {$2.239$};
\draw[->,thick] (pred2) -- (and) node [midway,right=2pt] {$2.322$};
\draw[->,thick] (noobssouth) -- (pred1) node [midway,left=2pt] {$1.077$};
\draw[->,thick] (tgtsouth) -- (pred2) node[midway,right=2pt]  {$1.052$};
\node at (0,2.5) {\vspace{0pt}};
\node at (0,-0.5) {\vspace{0pt}};
\end{tikzpicture}
\end{center}
\end{minipage}
\begin{minipage}{0.3\linewidth}
\includegraphics[width=0.99\linewidth]{blocksworld}
\end{minipage}}
\centerline{
\begin{minipage}{0.3\linewidth}
\scriptsize
\centerline{(a)}
\end{minipage}
\begin{minipage}{0.4\linewidth}
\scriptsize
\centerline{(b)}
\end{minipage}
\begin{minipage}{0.3\linewidth}
\scriptsize
\centerline{(c)}
\end{minipage}
}
\caption{Gridworld: (a) Avg. Rewards vs. Training Grids. (b) LNN Rule and (c) LNN-$\wedge$ for GoSouth(X,Y).}
\label{fig:gridworld}
\end{figure*}

Our experiments compare ILP with LNN against other neuro-symbolic ILP approaches on standard benchmarks. We evaluate rules in terms of application-specific goodness metrics and interpretability.

\subsection{Gridworld}

The goal in Gridworld is to learn rules that can help an agent move across an $N \times N$ regular grid. Some of the cells on the grid are deemed obstacles that the agent cannot step onto, and the agent's goal is to arrive at the cell which has been deemed the destination.

\noindent{\bf Predicates, Template and Rewards}: We include two kinds of base predicates to describe the grid 1) {\tt HasObstacle}-\emph{dir}$(X,Y)$ is {\tt true} if the cell next to $X,Y$ in direction \emph{dir} contains an obstacle, 2) {\tt HasTarget}-\emph{dir}$(X,Y)$ is {\tt true} if the destination lies in direction \emph{dir} from cell $X,Y$. There are four directions, North, South, East, and West, and including their negated counterparts, $\neg${\tt HasObstacle}-\emph{dir}$(X,Y)$ and $\neg${\tt HasTarget}-\emph{dir}$(X,Y)$, brings the total number of base predicates to $8$. The template is $S(X,Y) \leftarrow P(X,Y) \wedge Q(X,Y)$ where $P$'s domain includes all {\tt HasObstacle}-\emph{dir} predicates and their negated counterparts, and $Q$'s domain includes all {\tt HasTarget}-\emph{dir} predicates and their negated counterparts. We set $\alpha=0.8$ and use a simple reward mechanism: +1 for moving towards the destination, -1 for moving away, and -2 for stepping on an obstacle. The learning objective is to maximize rewards on randomly generated $5\times 5$ grids with $3$ obstacles sampled uniformly at random. We test the learned rules on grids with 12 obstacles. 

\noindent{\bf Results}: We compare our approach based on LNN-{\tt pred} and LNN-$\wedge$, against NeuralLP which uses attention and product $t$-norm. \myfigref{fig:gridworld} (a) shows mean rewards averaged across all cells of $50$ test grids produced by the learned rules on the y-axis vs. number of training grids observed on the x-axis. NeuralLP requires a lot more grids before it can learn the desired rules whereas we learn almost perfect rules after observing as few as $20$ training grids. Essentially, in comparison to product $t$-norm, due to the extra learnable parameters in LNN-$\wedge$, we can learn with fewer learning iterations. \myfigref{fig:gridworld} (b) shows the weights for the LNN rule for {\tt GoSouth}$(X,Y)$ on the edges and biases in parenthesis. This rule allows the agent to go South if 1) target is in that direction and, 2) there is no obstacle to the immediate South of the current cell. Despite the leaves of the template containing $8$ predicates each in their respective domains, the learned LNN-{\tt pred}s are quite sparse and thus highly interpretable. The left LNN-{\tt pred} assigns $1$ non-zero weight to $\neg${\tt HasObstacleSouth} out of all {\tt HasObstacle}-\emph{dir} predicates and their negated counterparts, and the right LNN-{\tt pred} assigns $1$ non-zero weight to {\tt HasTargetSouth} out of all {\tt HasTarget}-\emph{dir} predicates and their negated counterparts. This may be due to the use of {\tt relu1} in LNN-{\tt pred} whose sparsity-inducing properties have been noted before \citep{krizhevsky:cifar10}. In \myfigref{fig:gridworld} (c), we also plot  {\tt GoSouth}$(X,Y)$'s learned LNN-$\wedge$.

\subsection{Knowledge Base Completion}

Knowledge base completion (KBC) is a standard benchmark for ILP. We experiment with publicly available KBC datasets Kinship, UMLS \citep{kok2007statistical}, WN18RR \citep{dettmers2018convolutional}, and FB15K-237 \citep{toutanova2015observed} \footnote{All available at \url{github.com/shehzaadzd/MINERVA}} (see \mytabref{tab:stats} for statistics). We compare against a host of rule-based KBC approaches: NeuralLP \citep{yang:nips17}, DRUM \citep{sadeghian:neurips19}, CTP \citep{minervini:icml20} which is an improvement on neural theorem provers \citep{rocktaschel:nips17}, and the recently proposed RNNLogic\footnote{We compare with rules-only RNNLogic (``w/o embd."), since using embeddings is out of scope of this work.} \citep{qu:iclr21}.

\begin{table}
\caption{Statistics of KBC datasets.}
\centering
{\scriptsize
\begin{tabular}{rllll}
\toprule
Name & Vertices & Predicates & Facts & Queries\\
\hline
UMLS & 135 & 49 & 5216 & 661\\
Kinship & 104 & 26 & 10686 & 1074\\
WN18RR & 40945 & 11 & 86835 & 3134\\
FB15K-237 & 14505 & 237 & 272115 & 20466\\
\bottomrule
\end{tabular}}
\label{tab:stats}
\end{table}

\noindent{\bf Task Description and Template}: A popular abstraction of the KBC task is to complete edges or triples missing from the knowledge graph (KG). More precisely, given a query $\langle h, r, ? \rangle$, where $h$ denotes a source vertex and $r$ denotes a relation from the KG, most KBC approaches provide a ranked list of destination vertices. Most of the aforementioned rule-based KBC approaches exclusively focus on learning chain FOL rules as discussed in \mysecref{sec:relatedwork}. There are at least two prevalent approaches to learn chain rules for KBC. The first approach \citep{yang:nips17,sadeghian:neurips19} represents each predicate in the body as a mixture of relations present in the KG, and subsequently combines these via a conjunction operator. \myfigref{fig:example} (c)'s template captures this where the subtree rooted at $R$ defines chain rules of length $2$ predicates and $O$ captures length $1$ thus enabling learning of chain rules capturing multi-hop paths of length up to $2$. The only change we need to make to this template is to include all relations in the KG into the domains of the leaves.  It is also easy to extend the template to learn longer chain rules. A different approach, pioneered by MINERVA \citep{das:iclr18} and RNNLogic \citep{qu:iclr21}, is to define the chain rule as a mixture over all possible paths that can exist in the KG. This latter approach leads to more effective KBC and thus we report results by expressing it in our LNN framework as follows: 1) Express each possible multi-hop path as a \emph{base} relation, and 2) Use one LNN-pred operator to express a mixture over all multi-hop paths. 

\noindent{\bf Metrics and Methodology}: We learn a chain rule per relation present in the KG. Following previous work \citep{yang:nips17}, we also add inverse relations to the KG which switches the source and destination vertices. We also include inverse triples into our test set. We compute \emph{filtered} ranks for destinations \citep{bordes:nips13}, which removes all true triples ranked above, and compute the following metrics based on \citet{sun:acl20}'s suggestions. Let $n$ denote the number of destinations that have a score strictly greater than destination $t$'s and let the number of destinations assigned the same score as $t$'s be denoted by $m$ (including $t$), then we compute $t$'s mean reciprocal rank (MRR) and Hits@K as:  
\begin{align*}
\text{MRR} = \frac{1}{m} \sum_{r=n+1}^{n+m} \frac{1}{r}, \quad \text{Hits@K} = \frac{1}{m} \sum_{r=n+1}^{n+m} \delta(r \leq K)
\end{align*}
where $\delta()$ denotes the Dirac delta function. For each method, we report averages across all test set triples. We learn chain rules containing up to $3$ predicates for Kinship and UMLS, $4$ for FB15K-237, and $5$ for WN18RR in the body of the rule. We provide additional details including the training algorithm used and hyperparameter tuning in \myappref{app:details}.

\begin{table}
\caption{KBC Results: Bold font denotes best in row. CTP does not scale to WN18RR, FB15K-237. * indicates results copied from original paper.}
{\scriptsize
\centerline{
\begin{tabular}{crlllll}
\toprule
&& NeuralLP & DRUM & CTP & RNNLogic & Ours\\
\hline
\multirow{3}{*}{Kinship} & Hits@10 & 89.1 & 86.1 & 93.9 & 91.1 & {\bf 98.4}\\
& Hits@3 & 63.0 & 48.2 & 79.7 & 72.9 & \textbf{89.3}\\
& MRR & 48.8 & 40.0 & 70.3 & 64.5 & \textbf{81.9}\\
\hline
\multirow{3}{*}{UMLS} & Hits@10 & 93.0 & 97.9 & 97.0 & 91.1 & {\bf 99.4}\\
& Hits@3 & 75.4 & 91.2 & 91.0 & 82.1 & {\bf 98.3}\\
& MRR & 55.3 & 61.8 & 80.1 & 71.0 & {\bf 90.0}\\
\hline
\multirow{3}{*}{WN18RR} & Hits@10 & 50.2 & 52.1 & $-$ & 53.1$^*$ & {\bf 55.5}\\
& Hits@3 & 43.3 & 44.6 & $-$ & 47.5$^*$ & {\bf 49.7}\\
& MRR & 33.7 & 34.8 & $-$ & 45.5$^*$ & {\bf 47.3}\\
\hline
\multirow{3}{*}{FB15K-237} & Hits@10 & 32.8 & 33.1 & $-$ & 44.5$^*$ & \textbf{47.0}\\
& Hits@3 & 19.8 & 20.0 & $-$ & 31.5$^*$ & {\bf 34.2}\\
& MRR & 17.4 & 17.5 & $-$ & 28.8$^*$ & \textbf{30.7}\\
\bottomrule
\end{tabular}}}
\label{tab:kbcresults}
\end{table}


\noindent {\bf Results}: \mytabref{tab:kbcresults} reports results for all methods. While CTP improves upon the efficiency of neural theorem provers \citep{rocktaschel:nips17}, it still does not scale beyond Kinship and UMLS (indicated by $-$). Also, we copy previously published results for RNNLogic on WN18RR and FB15K-237\footnote{Despite exchanging multiple emails with its authors, we were unable to run RNNLogic code on the larger KBC datasets.} (indicated by $^*$). On the smaller datasets, CTP is the best baseline but our results are significantly better producing $16.5\%$ and $12.4\%$ relative improvements in MRR on Kinship and UMLS, respectively. On the larger datasets, RNNLogic is the current state-of-the-art within rule-based KBC and we outperform it producing $4\%$ and $6.6\%$ relative improvements in MRR on WN18RR and FB15K-237, respectively. Despite both learning a mixture over relation sequences appearing on KG paths, one reason for our relative success could be that RNNLogic uses an inexact training algorithm \citep{qu:iclr21}, relying on expectation-maximization, ELBO bound, whereas we employ no such approximations.

\noindent{\bf Learned Rules for FB15K-237}: \mytabref{tab:learnedrules} presents a few rules learned from FB15K-237. Rule 1 in \mytabref{tab:learnedrules} infers the language a person speaks by exploiting knowledge of the language spoken in her/his country of nationality. In terms of multi-hop path, this looks like: $P (\text{person}) \overset{\text{nationality}}{\longrightarrow} N (\text{nation}) \overset{\text{spoken\_in}}{\longrightarrow} L (\text{language})$. Similarly, Rule 2 uses the $\text{film\_country}$ relation instead of $\text{nationality}$ to infer the language used in a film. Besides $\text{spoken\_in}$, FB15K-237 contains other relations that can be utilized to infer language such as the $\text{official\_language}$ spoken in a country. Rule 3 uses this relation to infer the language spoken in a TV program by first exploiting knowledge of its country of origin. Rules 5, 6 and 7 are longer rules containing $3$ relations each in their body. Rule 5 infers a TV program's country by first exploiting knowledge of one of its actor's birth place and then determining which country the birth place belongs to. Rule 6 is similar but uses a film crew member's marriage location instead to infer the region where the film was released. Lastly, Rule 7 infers the marriage location of a celebrity by exploiting knowledge of where their friends got married.

\begin{table}
\caption{Learned rules from FB15K-237.}
{\scriptsize
$\arraycolsep=1pt\begin{array}{crl}
1) & \text{person\_language}(P,L) \leftarrow~& \text{nationality}(P,N) \land~ \text{spoken\_in}(L,N)\\
2) & \text{film\_language}(F,L) \leftarrow~& \text{film\_country}(F,C) \land~ \text{spoken\_in}(L,C)\\
3) & \text{tv\_program\_language}(P,L) \leftarrow~& \text{country\_of\_tv\_program}(P,N) ~\land\\
&& \text{official\_language}(N,L)\\
4) & \text{burial\_place}(P,L) \leftarrow~& \text{nationality}(P,N) \land~ \text{located\_in}(L,N)\\
5) & \text{tv\_program\_country}(P,N) \leftarrow~& \text{tv\_program\_actor}(P,A) ~\land\\
&& \text{born\_in}(A, L) \land~ \text{located\_in}(L,N)\\
6) & \text{film\_release\_region}(F,R) \leftarrow~& \text{film\_crew}(F,P) ~\land\\
&& \text{marriage\_location}(P,L) \land~ \text{located\_in}(L,R)\\
7) & \text{marriage\_location}(P,L) \leftarrow~& \text{celebrity\_friends}(P,F) ~\land\\
&& \text{marriage\_location}(F,L^\prime) \land~ \text{location\_adjoins}(L^\prime,L)
\end{array}$
}
\label{tab:learnedrules}
\end{table}

\noindent{\bf Additional KBC Results}: Due to space constraints, in \myappref{app:countries} we report results on the Countries dataset \citep{bouchard:aaai15} for which ground truth rules are known. On Countries, our KBC accuracy is comparable to other approaches and the learned LNN rules form a close match with the ground truth rules specified in \citet{nickel:aaai16}.

\section{Conclusion}
\label{sec:conclusion}


Our experiments show that learning rules and logical connectives jointly is not only possible but leads to more accurate rules than other neuro-symbolic ILP approaches. Templates provide a flexible way to express a wide range of ILP tasks. The templates used for Gridworld and KBC are distinct, yet we outperformed baselines in both cases. LNN rules use weights sparingly and are eminently interpretable while LNN operators' constraint formulation ensures close ties to classical logic's precise semantics compared to other approaches (e.g., NLM). While our neural network requires grounding the KB, our approach is still scalable enough to tackle the larger KBC benchmarks whereas others are not (e.g., CTP). In terms of future work, we aim to combine the ideas presented here with embedding of predicates and constants in a high-dimensional latent space to hopefully further improve performance. We would also like to extend our approach to learn more general Prolog-style rules.

\bibliography{lnn-ilp}

\appendix

\section{Implementation Details for KBC Experiments}
\label{app:details}

Given a training knowledge graph $\gG = \langle \gV, \gR, \gE \rangle$ whose vertices are given by $\gV$, edges are given by $\gE$ and relations are given by $\gR$, we learn a chain FOL rule for each $r \in \gR$ independently. Each edge in $\gE$ is given by a triple $\langle h, r, t \rangle$ where $h, t \in \gV$ denote source and destination vertices, and $r \in \gR$ denotes a relation. Following previous work \citep{yang:nips17}, we also add inverse relations. In other words, for each $r \in \gR$ we introduce a new relation $r^{-1}$ by adding for each $\langle h, r, t \rangle \in \gE$ a new triple $\langle t, r^{-1}, h \rangle$ to $\gG$. Learning to predict destinations for a given relation $r$ in the augmented set of relations is essentially a binary classification task where $\langle h, r, t \rangle \in \gE$ and $\langle h, r, t \rangle \notin \gE, \forall h, t \in \gV$, denote positive and negative examples, respectively.

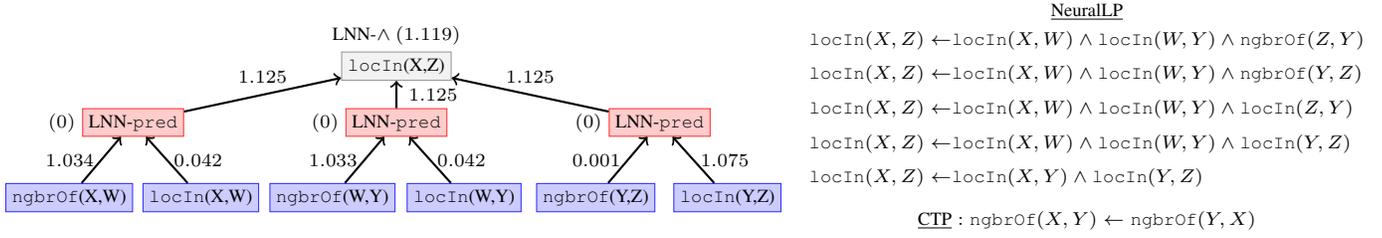
\begin{figure*}
\begin{minipage}{0.58\linewidth}
\scriptsize
\begin{tikzpicture}[=>latex]
\node (pred11) at (0,0) [draw=blue!80,fill=blue!20] {{\tt ngbrOf}(X,W)};
\node (pred12) at (1.75,0) [draw=blue!80,fill=blue!20] {{\tt locIn}(X,W)};
\node[label=left:{$(0)$}] (pred1) at (0.85,1) [draw=red!80,fill=red!20] {LNN-{\tt pred}};
\draw[->,thick] (pred11) -- (pred1) node[midway,left]{$1.034$};
\draw[->,thick] (pred12) -- (pred1) node[midway,right]{$0.042$};
\node (pred21) at (3.5,0) [draw=blue!80,fill=blue!20] {{\tt ngbrOf}(W,Y)};
\node (pred22) at (5.25,0) [draw=blue!80,fill=blue!20] {{\tt locIn}(W,Y)};
\node[label=left:{$(0)$}] (pred2) at (4.35,1) [draw=red!80,fill=red!20] {LNN-{\tt pred}};
\draw[->,thick] (pred21) -- (pred2) node[midway,left]{$1.033$};
\draw[->,thick] (pred22) -- (pred2) node[midway,right]{$0.042$};
\node (pred31) at (7,0) [draw=blue!80,fill=blue!20] {{\tt ngbrOf}(Y,Z)};
\node (pred32) at (8.75,0) [draw=blue!80,fill=blue!20] {{\tt locIn}(Y,Z)};
\node[label=left:{$(0)$}] (pred3) at (7.85,1) [draw=red!80,fill=red!20] {LNN-{\tt pred}};
\draw[->,thick] (pred31) -- (pred3) node[midway,left]{$0.001$};
\draw[->,thick] (pred32) -- (pred3) node[midway,right]{$1.075$};
\node[label=above:{LNN-$\wedge$ $(1.119)$}] (and) at (4.35,1.75) [draw=gray!80,fill=lightgray!20] {{\tt locIn}(X,Z)};
\draw[->,thick] (pred1) -- (and) node[midway,above=2pt] {$1.125$};
\draw[->,thick] (pred2) -- (and) node[midway,right=2pt] {$1.125$};
\draw[->,thick] (pred3) -- (and) node[midway,above=2pt] {$1.125$};
\end{tikzpicture}
\end{minipage}
\begin{minipage}{0.45\linewidth}
\scriptsize
\centerline{\underline{NeuralLP}}
\vspace{-16pt}
\begin{align*}
\text{{\tt locIn}}(X, Z) \leftarrow &\text{{\tt locIn}}(X, W) \wedge \text{{\tt locIn}}(W, Y) \wedge \text{{\tt ngbrOf}}(Z, Y) \\
\text{{\tt locIn}}(X, Z) \leftarrow &\text{{\tt locIn}}(X, W) \wedge \text{{\tt locIn}}(W, Y) \wedge \text{{\tt ngbrOf}}(Y, Z)\\
\text{{\tt locIn}}(X, Z) \leftarrow &\text{{\tt locIn}}(X, W) \wedge \text{{\tt locIn}}(W, Y)\wedge \text{{\tt locIn}}(Z, Y)\\
\text{{\tt locIn}}(X, Z) \leftarrow &\text{{\tt locIn}}(X, W) \wedge \text{{\tt locIn}}(W, Y) \wedge \text{{\tt locIn}}(Y, Z)\\
\text{{\tt locIn}}(X, Z) \leftarrow & \text{{\tt locIn}}(X, Y) \wedge \text{{\tt locIn}}(Y, Z)\\
\end{align*}
\vspace{-24pt}
\begin{equation*}
\text{\underline{CTP}}: \text{{\tt ngbrOf}}(X, Y) \leftarrow \text{{\tt ngbrOf}}(Y, X)
\end{equation*}
\end{minipage}
\caption{LNN-rule (left) vs. NeuralLP's (5) rules (top right) vs. CTP's rule (bottom right) for Countries-S3.}
\label{fig:countries}
\end{figure*}

\noindent{\bf Training Algorithm \& Hyperparameter Settings}: In each iteration, we sample uniformly at random a mini-batch of positive triples $B^+$ from $\langle h,r,t \rangle \in \gE$ and negative triples $B^-$ from $\langle h,r,t \rangle \notin \gE, \forall h,t \in \gV$, such that $|B^+| = |B^-|$ to minimize the following loss:
\begin{equation*}
\sum_{\langle h, r, t \rangle \in B^+} \sum_{\langle h^\prime, r, t^\prime \rangle \in B^-} \max\{0, \text{score}(h^\prime,t^\prime) - \text{score}(h,t) + \gamma\}
\end{equation*}
where $\gamma$ denotes the margin hyperparameter. We use Adagrad \citep{duchi:jmlr11} with step size $\in \{0.1, 1.0\}$, margin $\gamma \in \{0.1, 0.5, 1.0, 2.0\}$ and batch size $|B^+| = |B^-| = 8$. We use the validation set to perform hyperparameter tuning and learn rules of length up to $4$ for FB15K-237, $5$ for WN18RR, and $3$ for Kinship, UMLS. 

\section{Additional KBC Experiments}
\label{app:countries}

Besides the experiments presented in \mysecref{sec:experiments}, we also experiment with the \emph{Countries} dataset \citep{bouchard:aaai15} which contains $272$ vertices, $1158$ facts and $2$ predicates {\tt locatedIn}$(X,Y)$ (abbreviated {\tt locIn}) and {\tt neighborOf}$(X,Y)$ (abbreviated {\tt ngbrOf}). Following previous work \citep{das:iclr18}, we report area under the precision-recall curve (AUC-PR). 

\begin{table}
\caption{AUC-PR Results on Countries}
{\scriptsize
\centerline{
\begin{tabular}{c|ccccc}
\toprule
& NLM & NTP-$\lambda$ & CTP & NeuralLP & Ours \\
\hline
S1 & { 58.06 $\pm$ 2.4} & {\bf 100 $\pm$ 0} & 99.6 $\pm$ 0.5 & {\bf 100 $\pm$ 0} & {\bf 100 $\pm$ 0}\\
S2 & { 40.57 $\pm$ 5.3} &{\bf 93.04 $\pm$ 0.4} & 92.4 $\pm$ 1.7 & 75.1 $\pm$ 0.3 & 92.3 $\pm$ 0\\
S3 & { 53.37 $\pm$ 2.8} &77.26 $\pm$ 17.0 & 55.4 $\pm$ 3.5 & {\bf 92.2 $\pm$ 0.2} & 91.3 $\pm$ 0\\
\bottomrule
\end{tabular}}}
\label{tab:countries}
\end{table}

\citet{nickel:aaai16} permute the facts in this dataset to pose $3$ learning tasks $\{S1, S2, S3\}$ each corresponding to learning a different rule. S1 and S2's rules contain $2$ predicates in their bodies, whereas S3's body contains $3$. We use $S(X,Z) \leftarrow P(X,Y) \wedge Q(Y,Z)$ as template for S1 and S2, and $S(X,Z) \leftarrow P(X,W) \wedge Q(W,Y) \wedge O(Y,Z)$ for S3, where $\text{Dom}(P) = \text{Dom}(Q) = \text{Dom}(O) = \{\text{{\tt locIn}}, \text{{\tt ngbrOf}}\}$. We compare against NeuralLP\footnote{\url{github.com/fanyangxyz/Neural-LP}} \citep{yang:nips17}, neural theorem provers\footnote{\url{github.com/uclnlp/ntp}} (NTP-$\lambda$) \citep{rocktaschel:nips17}, conditional theorem provers\footnote{\url{github.com/uclnlp/ctp}} (CTP) \citep{minervini:icml20} and neural logic machines\footnote{\url{github.com/google/neural-logic-machines}} (NLM) \citep{dong:iclr19}. 

\mytabref{tab:countries} reports averages across 3 runs. The NLM implementation we report results with is quite pessimistic almost never predicting a link. All other approaches achieve perfect AUC-PR on S1. We outperform NeuralLP on S2 and NTP shows high variance on S3\footnote{NTP's high variance on S2 is also noted in \citet{das:iclr18}.} perhaps due to its excessive parameterization (NTP learns embeddings for predicates \emph{and} vertices). CTP performs well on the first two tasks but fails on S3. To find out why, we take a close look at the learned rules next.

The goal in S2 and S3 (Countries) is to learn the following rules \citep{nickel:aaai16}:
\squishlist
\item[] (S2) {\tt locIn}(X,Z) $\leftarrow$ {\tt ngbrOf}(X,Y) $\wedge$ {\tt locIn}(Y,Z)
\item[] (S3) {\tt locIn}(X,Z) $\leftarrow$ {\tt ngbrOf}(X,W) $\wedge$ {\tt ngbrOf}(W,Y) $\wedge$ {\tt locIn}(Y,Z)
\squishend

We compare the rules learned by our approach, NeuralLP and CTP for Countries-S3. NeuralLP produces a weighted list of $5$ rules (see \myfigref{fig:countries} top right, weights omitted) none of which capture the correct rule shown above. CTP learns that {\tt ngbrOf} is a symmetric relation \citep{minervini:icml20} which makes sense (\myfigref{fig:countries} bottom right). Unfortunately however, all facts that need to be proved in the test set of Countries-S3 pertain to {\tt locIn} predicate which explains why CTP's AUC-PR for Countries-S3 is so poor (\mytabref{tab:countries}). In contrast, the learned LNN-rule shown in \myfigref{fig:countries} (left) is a near perfect translation of the correct logical rule into real-valued logic. Note that, the leftmost and middle LNN-{\tt pred}s place a large weight on {\tt ngbrOf} vs. a small weight on {\tt locIn} while the rightmost LNN-{\tt pred} does the opposite, which lines up perfectly with the rule to be learned.

\begin{figure}[b]
\scriptsize
\centering
\begin{tikzpicture}[=>latex]
\node (pred11) at (0,0) [draw=blue!80,fill=blue!20] {{\tt ngbrOf}(X,Y)};
\node (pred12) at (1.75,0) [draw=blue!80,fill=blue!20] {{\tt locIn}(X,Y)};
\node[label=left:{$(0)$}] (pred1) at (0.85,1) [draw=red!80,fill=red!20] {LNN-{\tt pred}};
\draw[->,thick] (pred11) -- (pred1) node[midway,left]{$1.228$};
\draw[->,thick] (pred12) -- (pred1) node[midway,right]{$0.002$};
\node (pred21) at (3.5,0) [draw=blue!80,fill=blue!20] {{\tt ngbrOf}(Y,Z)};
\node (pred22) at (5.25,0) [draw=blue!80,fill=blue!20] {{\tt locIn}(Y,Z)};
\node[label=right:{$(0)$}] (pred2) at (4.35,1) [draw=red!80,fill=red!20] {LNN-{\tt pred}};
\draw[->,thick] (pred21) -- (pred2) node[midway,left]{$0.04$};
\draw[->,thick] (pred22) -- (pred2) node[midway,right]{$1.186$};
\node[label=above:{LNN-$\wedge$ $(1.056)$}] (and) at (2.6,1.75) [draw=gray!80,fill=lightgray!20] {{\tt locIn}(X,Z)};
\draw[->,thick] (pred1) -- (and) node[midway,left=2pt] {$1.059$};
\draw[->,thick] (pred2) -- (and) node[midway,right=2pt] {$1.059$};
\end{tikzpicture}
\caption{LNN rule learned from Countries-S2}
\label{fig:countriess2}
\end{figure}
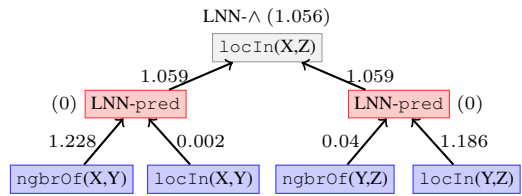

We also compared the rules learned for S2. Just as in the case of S3, for S2, the learned LNN rule's (shown in \myfigref{fig:countriess2}) left LNN-{\tt pred} places a large weight on {\tt ngbrOf} while placing a small weight on {\tt locIn} whereas the right LNN-{\tt pred} does the opposite, closely matching the correct FOL rule shown above. The list of rules learned by NeuralLP (weight depicted in parenthesis) is shown below:

\centerline{\begin{minipage}{0.9\linewidth}
\begin{align*}
\text{{\tt locIn}}(X, Z) &\leftarrow \text{{\tt locIn}}(X, Y) \wedge \text{{\tt ngbrOf}}(Z, Y)\\
\text{{\tt locIn}}(X, Z) &\leftarrow \text{{\tt locIn}}(X, Y) \wedge \text{{\tt ngbrOf}}(Y, Z)\\
\text{{\tt locIn}}(X, Z) &\leftarrow \text{{\tt locIn}}(X, Y) \wedge \text{{\tt locIn}}(Z, Y)\\
\text{{\tt locIn}}(X, Z) &\leftarrow  \text{{\tt locIn}}(X, Z)\\
\text{{\tt locIn}}(X, Z) &\leftarrow  \text{{\tt locIn}}(X, Y) \wedge \text{{\tt locIn}}(Y, Z)
\end{align*}
\end{minipage}}

\noindent none of which match the correct rule to be learned for S2. CTP learns the following rule \citep{minervini:icml20}:\\
\centerline{\begin{minipage}{0.9\linewidth}
\begin{align*}
\text{{\tt ngbrOf}}(X, Z) \leftarrow \text{{\tt ngbrOf}}(X,Y) \wedge \text{{\tt locIn}}(Y,Z)\\
\end{align*}
\end{minipage}}
\noindent Not only does this rule not match the correct rule to be learned, it is also not very useful from the perspective of proving facts in the test set which pertain to predicate {\tt locIn} exclusively.

\end{document}